\pdfoutput=1
\documentclass{bmvc2k}

\newcommand\blfootnote[1]{%
  \begingroup
  \renewcommand\thefootnote{}\footnote{#1}%
  \addtocounter{footnote}{-1}%
  \endgroup
}
\title{DeepUME: Learning the Universal Manifold Embedding for Robust Point Cloud Registration}

\addauthor{Natalie Lang}{langn@post.bgu.ac.il}{0}
\addauthor{Joseph  M. Francos}{francos@ee.bgu.ac.il}{0}

\addinstitution{Ben-Gurion University\\
 Beer-Sheva, Israel
}

\runninghead{Lang, Francos}{DeepUME}

\usepackage{booktabs, multirow}
\usepackage{comment}
\usepackage{enumitem}
\usepackage{amsmath}
\usepackage{amsfonts}
\usepackage{amssymb}
\usepackage{mathtools}
\usepackage{makecell}
\usepackage{xcolor}
\setlist[itemize]{itemsep=0pt, topsep=0pt}
\usepackage[belowskip=-10pt,aboveskip=0pt]{caption}
\DeclareCaptionFormat{myformat}{\fontsize{7}{8}\selectfont#1#2#3}
\usepackage{subfiles} 

\newcommand{\R}{\mathbb{R}}

\DeclareMathOperator{\Mom}{\mathbf{M}}

\newcommand{\cO}{\mathcal{O}}
\newcommand{\cF}{\mathcal{F}}

\newcommand{\cN}{\mathcal{N}}

\newcommand{\cC}{\mathcal{C}}
\newcommand{\cPC}{\mathcal{P}}

\newcommand{\SOG}{\text{SO}}
\newcommand{\UME}{\text{UME}}

\newcommand{\norm}[1]{{\lVert#1\rVert}_2}
\newcommand{\mathset}[1]{\left\{#1\right\}}

\DeclarePairedDelimiter\abs{\lvert}{\rvert}

\newcommand{\vx}{\mathbf{x}}
\newcommand{\vm}{\mathbf{m}}
\newcommand{\vc}{\mathbf{c}}
\newcommand{\vv}{\mathbf{v}}
\newcommand{\vu}{\mathbf{u}}
\newcommand{\vt}{\mathbf{t}}
\newcommand{\vp}{\mathbf{p}}
\newcommand{\vR}{\mathbf{R}}
\newcommand{\vD}{\mathbf{D}}
\newcommand{\vA}{\mathbf{A}}
\newcommand{\vT}{\mathbf{T}}

\def\ie{\emph{i.e}\bmvaOneDot}
\def\eg{\emph{e.g}\bmvaOneDot}

\def\etal{\emph{et al}\bmvaOneDot}

\newtheorem{theorem}{Theorem}[section]
\begin{document}
\maketitle
\blfootnote{This research was supported by NSF-BSF Computing and Communication Foundations (CCF) grants, CCF-2016667, and BSF-2016667 and by the Israeli Ministry of Innovation, Science and Technology grant 3-16583.}
\vspace{-4mm}
\begin{abstract}
Registration of point clouds related by rigid transformations is one of the fundamental problems in computer vision. 
However, a solution to the practical scenario of aligning sparsely and differently sampled observations in the presence of noise  is still lacking. 
We approach registration in this scenario with a fusion of the closed-form Universal Manifold Embedding (UME) method and a deep neural network. 
The two are combined into a single unified framework, named DeepUME, trained end-to-end and in an unsupervised manner. 
To successfully provide a global solution in the presence of large transformations, we employ an $\SOG(3)$-invariant coordinate system to learn both a  joint-resampling strategy of the point clouds and $\SOG(3)$-invariant features. These features are then utilized by the geometric UME method for transformation estimation.
The parameters of DeepUME are optimized using a metric designed to overcome an ambiguity problem emerging in the registration of symmetric shapes, when noisy scenarios are considered. We show that our hybrid method outperforms state-of-the-art registration methods in various scenarios, and generalizes well to unseen data sets. Our code is publicly available\footnote{\url{https://github.com/langnatalie/DeepUME}}.
\end{abstract}
\vspace{-4mm}
\begin{figure*}[hbt!] 
\centering
\includegraphics[width=0.9\textwidth]{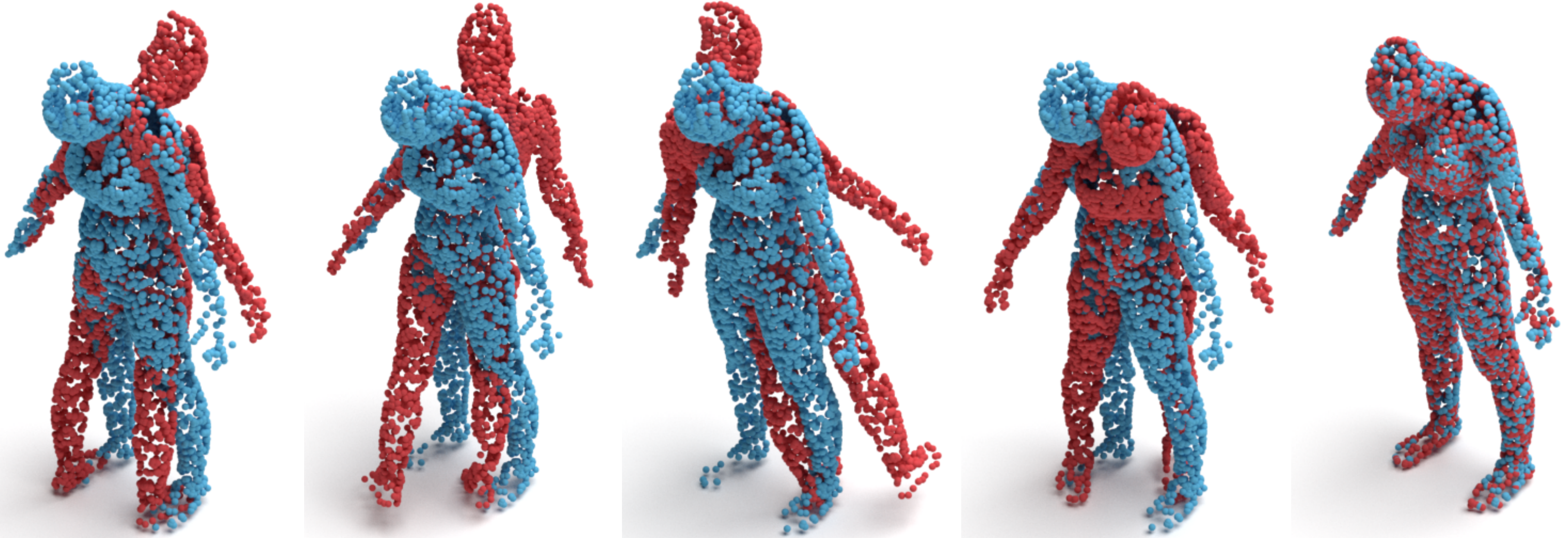}
   \caption*{{ 
   {\color{white} AAA}
   Input 
   {\color{white} AAAAAAAAA}
   UME \cite{efraim2019universal} 
   {\color{white} AAAAAAAA}
   DCP \cite{wang2019deep} 
   {\color{white} AAAAAA}
   DeepGMR \cite{yuan2020deepgmr} 
   {\color{white} AAA}
   DeepUME (ours) 
   {\color{white} }}}
   \vspace{4mm}
   \captionsetup{width=0.9\textwidth}
   \caption{
   Registration results on an unseen data set, where the observations are subject to a large relative rotation and sampling noise (zero-intersection model).
    While UME \cite{efraim2019universal} and DCP \cite{wang2019deep} fail to align the objects, and DeepGMR \cite{yuan2020deepgmr} results with a substantial registration error, the proposed method successfully aligns the shapes.}
\label{fig:teaser}
\end{figure*}

\vspace{-4mm}
\section{Introduction}\label{sec:intro}
\vspace{-3mm}
The massive development of 3D range sensors \cite{collis1970lidar,smisek20133d} led to an intense  interest in 3D data analysis. As 3D data is commonly acquired in the form of a point cloud, many related applications have been studied in recent years for that data form.
In wide range of applications, specifically in medical imaging \cite{hill2001medical}, autonomous driving \cite{bresson2017simultaneous} and robotics \cite{durrant2006simultaneous}, the alignment of 3D objects into a coherent world model is a crucial problem.
Point cloud rigid alignment is a deep-rooted problem in computer vision and graphics, and various methods for point cloud registration have been suggested \cite{pomerleau2015review}.


In general, the point clouds to be registered are sampled from a physical object. When two point clouds are sampled at two different poses of an object, and especially when sampling is sparse, it is unlikely that the same set of object points is sampled in both. The difference between the sampling patterns of the object may result in model mismatch when performing registration,  and we therefore refer to it as sampling noise. 
Registration of point clouds in the presence of noise has been extensively studied by both closed-form \cite{besl1992method,yang2015go,zhou2016fast,rusinkiewicz2001efficient} and learning-based \cite{aoki2019pointnetlk,wang2019deep,yuan2020deepgmr,Fu2021RGM} methods. In most of these works, the noise is modeled as an Additive White Gaussian Noise (AWGN) on the coordinates. However, in many registration applications the point clouds are sampled differently and sparsely. In such applications,
these sampling effects are dominant and adversely   affect registration performance, yet they cannot be modeled by an AWGN.


In this work we address the global registration of 3D under-sampled point clouds, where the point clouds are differently sampled, and the samples are subject to the presence of an additive coordinate noise. 
Our strategy is to combine the closed-form  Universal Manifold Embedding (UME) registration method \cite{efraim2019universal}, and a learning-based framework.
The UME non-linearly maps functions related by geometric transformations of coordinates (rigid, in our case) to matrices that are linearly related by the transformation parameters. In the UME framework, the embedding of the orbit of possible observations on the object to the space of matrices is based on constructing an operator that evaluates a sequence of low-order geometric moments of some function defined on the point clouds to be registered. This representation is therefore more resilient to noise than local operators, as under reasonable noise, the geometric structure of the point cloud is preserved. Since the UME is an operator defined on functions of the coordinates, in order to enable registration, these functions (features) need to be invariant to the transformation. 
While in the original UME framework, the invariant features are\textit{ hand-crafted functions}, in this work we \textit{learn} those from data using an unsupervised deep neural network architecture.

The proposed framework is a  cascade of three blocks:
The first is a pre-processing step that employs an SO(3)-invariant coordinate system, constructed using PCA, to enable estimation of large transformations.
The second block is the neural network,  designed to implement  joint-resampling  and embedding of the raw point clouds.
The final block implements the UME. 
Our trained model is tested on both seen and unseen data sets, to demonstrate generalization capabilities. Inference performance is evaluated for different noise scenarios using metrics that are invariant to the ambiguity arising in symmetric shapes registration when 
noisy scenarios are considered. 

Our main contributions are as follows:
\begin{itemize}
 \item We integrate, for the first time, the closed-form UME registration methodology and a data-driven approach by both adapting the UME method to the DNN framework and designing the DNN architecture to optimize the UME performance. We address the highly practical yet less studied case of registering point clouds that are sparsely and randomly sampled in the presence of large transformations (full range of rotations).
 Our hybrid model is trained end-to-end, labels free and results with substantial performance gains compared to  competing state-of-the-art methods. 

\item To enhance the DeepUME performance in the presence of large deformations and since the point clouds are sparsely and differently sampled, we present a learned joint resampling
of both point clouds to be registered, using a novel approach for integrating  a PCA module, a Transformer module and a DGCNN module. By mapping the input data to an $\SOG(3)$-invariant coordinates system, we overcome DGCNN inability to learn invariant features for registration in the case of large rotations. 
The Transformer is used for implementing a joint-resampling strategy of both point clouds to be registered. However, while learned joint-resampling procedures are usually applied in a \textit{high-dimensional} feature space, in our framework, the joint-resampling is aimed at "equalizing" the differences in the sampling patterns of the observations, and is therefore implemented in the coordinate \textit{low-dimensional} space.
\end{itemize}

\vspace{-4mm}
\section{Related Work}
\vspace{-3mm}

There are many approaches to 3D point cloud registration. One of the commonly practiced approaches is to extract and match spatially local features \eg,  \cite{Guo2016,SPINIM,RUSU2008,Yang2016,Yang2017,LRF}.
Many of the existing methods are 3D adaptations of 2D image processing solutions, such as variants of 3D-SIFT \cite{3DSIFT} and the 3D Harris key-point detector \cite{Harris3D}.
In 3D, with the absence of a regular sampling grid, artifacts, and sampling noise, key-point matching is prone to high outlier rates and localization errors.
Hence, global alignment  estimated by key-point matching usually employs outlier rejection methods such as RANSAC \cite{fischler1981random} followed by a refinement stage using local optimization algorithms \cite{ICP1,Zhang1994,POTTMANN2004,3DNDT}. DGR \cite{Choy_2020_CVPR} follows a similar paradigm, but inlier detection is learnable.
Numerous works have been proposed for handling outliers and noise \cite{chetverikov2005robust}, formulating robust minimizers \cite{fitzgibbon2003robust}, or proposing more suitable distance metrics.
The standard algorithm in the category of refinement algorithms, also known as local registration, is the Iterative Closest Point algorithm (ICP) \cite{ICP1,Zhang1994}. It constructs point correspondences based on spatial proximity followed by a transformation estimation step. Over the years, many variants of the ICP algorithm have been proposed in attempt to improve the convergence rate, robustness, and accuracy of the algorithm.

Registration methods are not restricted only to methods based on the extraction and matching of key-points. In \cite{isola2011makes}, for example, an initial alignment is found by employing a matched filter in the frequency space of local orientation histograms. In \cite{Shimshoni} an initial alignment is found by clustering the orientations of local point cloud descriptors followed by estimating the relative rotation between clusters. In this work, a global closed-form solution that employs the UME \cite{efraim2019universal} representation of the shapes to be registered, is being integrated. As a result, an efficient and accurate registration scheme is achieved where no initial alignment is required.


Other types of registration methods adopt learning-based techniques. Pioneered by PointNet \cite{qi2017pointnet} and subsequently by DGCNN \cite{wang2019dynamic}, point cloud representations are learned from data in a task-specific manner. These can, in turn, be leveraged for robust point cloud registration \eg, \cite{qi2017pointnet++,su2018splatnet,wang2019deep, wang2019prnet}. 
PointNetLK \cite{aoki2019pointnetlk} minimizes learned feature distance by a differentiable Lucas-Kanade algorithm \cite{lucas1981iterative}.
DCP \cite{wang2019deep} addressed feature matching via attention-based module and differentiable SVD modules for point-to-point registration.
The recently proposed RGM \cite{Fu2021RGM} transforms point clouds into graphs and perform deep graph matching for extracting deep features soft correspondence matrix.
In section \ref{sec:Experiments} we show that registration performance using the aforementioned architectures deteriorates when observations are related by large rotations. 
We relax that limitation by employing an $\SOG$-(3)invariant coordinate system, and therefore provide a global registration solution.
DeepGMR \cite{yuan2020deepgmr} also addresses this limitation by extracting pose-invariant correspondences between raw point clouds and Gaussian mixture model (GMM) parameters, and recovers the transformation from the matched mixtures. However, its performance deteriorates in the presence of sampling noise.

\vspace{-4mm}
\section{Problem Definition}
\vspace{-3mm}
A point cloud $\cPC$ is a finite set of points in  $\R^3$. Usually, these points are samples  from a physical object, $\cO\subseteq \R^3$ (we may think of it as a surface or a manifold). Viewing point clouds as sets of samples, the registration problem may be formulated as follows: Let $\cO\subseteq \R^3$ be a physical object and $T(\vx)=\vR\vx+\vt$ a rigid map  ($\vR\in \SOG(3)$ is a rotation matrix and $\vt\in \R^3$ is translation vector). We consider the transformed object 
$T(\cO):= \{ T(\vx) ~:~ \vx\in \cO \}.$
Let $\cPC_1$ and $\cPC_2$ be two point clouds sampled from the object $\cO$ and the transformed object $T(\cO)$, respectively. In the registration problem, the objective is to estimate the transformation parameters $\vR$ and $\vt$ given only $\cPC_1$ and $\cPC_2$. 

Since point clouds are generated by a sampling procedure, the effects of sampling must be addressed, when solving the registration problem. Ideally, the relation between the two sampled point clouds $\cPC_1$ and $\cPC_2$ (sampled from $\cO$ and $T(\cO)$, respectively) satisfies the relation $\cPC_2=T(\cPC_1)$. Unfortunately, 
when point clouds are sparsely sampled at two different poses of some object,  it is unlikely that the same set of object points is sampled, in both: 
If we assume a uniformly distributed sampling pattern on a continuous surface, it may be easily proved that the probability of having such a relation is null. 
As we show in our experiments, once sparse and differently-sampled point clouds are considered, the sampling differences result in substantial registration errors, having different characteristics from those of AWGN on the coordinates. 

Under-sampled point cloud registration is considered in the following scenarios: 
\begin{itemize}
    \item Full intersection (Vanilla model) - where $\cPC_2=T(\cPC_1)$.
    \item Sampling noise - Two cases are considered: Partial intersection, where $\cPC_2$ and  $T(\cPC_1)$ may intersect, but are not identical; Zero intersection, where $\cPC_2$ and $T(\cPC_1)$ have no samples in common.
    \item Gaussian noise - $\cPC_2=T(\cPC_1)+\mathcal{N}$, where  $\cPC_2$ is a result of a rigid transformation of $\cPC_1$ with its coordinates perturbed by AWGN.
\end{itemize}
\vspace{-4mm}
\section{DeepUME}\label{sec:DeepUME}
\vspace{-3mm}
\begin{figure*}
\centering
\includegraphics[width=\textwidth]{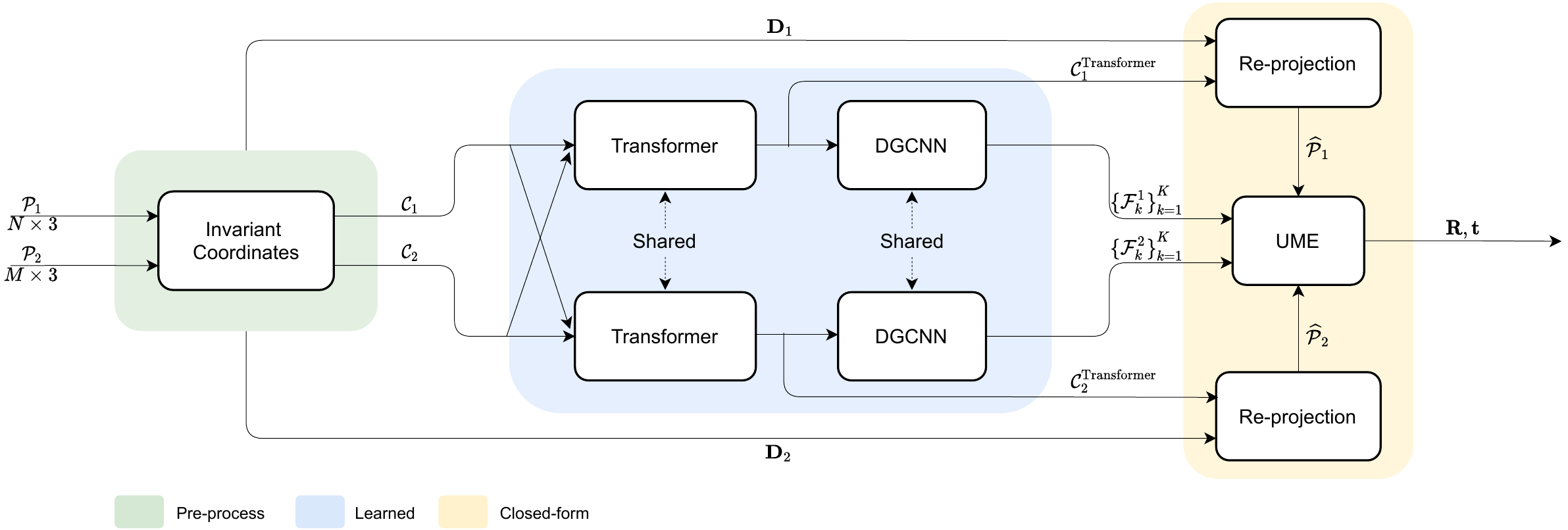}
\vspace{-3mm}
\caption{DeepUME network architecture.}
\label{fig:arcitecture}
\end{figure*}
Following the strategy of integrating the UME registration methodology into a deep neural network, we both adjust the UME method \cite{efraim2019universal} adapting it to a DNN framework and design our architecture to optimize the UME performance. 
The proposed framework is a  cascade of three blocks:
The first is a pre-processing step that employs an SO(3)-invariant coordinate system, constructed using PCA, that enables estimation of large transformations.
The second block is the neural network,  designed to implement  joint-resampling  and embedding of the raw point clouds aimed at "equalizing"  the differences in the  sampling patterns of the observations, and consequently boost the performance of the
third block, implementing the UME. 
The framework is illustrated in Figure \ref{fig:arcitecture} (see supplementary for detailed illustrations). The details about its building blocks, and in particular  the Transformer and DGCNN architectures adaptations into the UME framework are provided in the following. 

In the UME registration framework the input is composed of two point clouds satisfying the relation $\cPC_2=\vR\cdot \cPC_1 + \vt$, and an invariant feature (that is, a function $\cF$ defined on a point cloud $\cPC$ such that $\cF(\vp)=\cF(\vR\cdot \vp+\vt)$ for any $\vp\in \cPC$). Applying the UME operator on each of the point clouds, two matrices $\Mom_{\cPC_1}$ and $\Mom_{\cPC_2}$ are obtained, such that $\Mom_{\cPC_2} = \vR \cdot \Mom_{\cPC_1}$.   
The geometric nature of the UME motivates us to use it as a basis for our framework. While many registration methods find corresponding points in the reference and the transformed point clouds in order to solve the registration problem, in the UME methodology a new set of "corresponding points" is constructed by evaluating low order geometric moments of the invariant feature. This is a significant advantage in noisy scenarios, where  point correspondences between the reference and transformed point clouds, may not exist at all. The use of moments allows us to exploit the geometric structure of the objects to be registered, which is invariant under sampling (as long as the sampling is reliable) resulting in an improved immunity to sampling noise.

The goal of the deep neural network in our framework is to construct multiple high-quality invariant features, in order to maximize the performance of the UME in various noisy scenarios. We adopt the joint usage of DGCNN and Transformer blocks \cite{wang2019deep}, which has been proven to be very efficient in creating high-dimensional embedding  for point clouds, and adapt it to the UME framework in order to \textit{learn} multiple high-quality $\SOG(3) $ invariant features.

\textbf{UME registration} 
The UME framework \cite{itmanifolds,efraim2019universal} is designed for registering two functions $f,g:\R^n\to \R$, with compact supports related by a geometric transformation (rigid, affine) parameterized by $\vA$. Zero and first order moments (integrals) are evaluated in constructing the UME matrix of dimension $(n+1)\times D$ (where $D> n+1$). The UME matrices of $f$ and $g$ satisfy the relation: $\UME_f = \vA\cdot\UME_g$.
Following the principles of the UME, we derive a new discrete closed-form implementation of the UME for the registration of point clouds undergoing rigid transformations. 

More specifically, let $\cPC_1$ and $\cPC_2$ be two point clouds related by a rigid transformation $\vT$. Then, an invariant feature (function) $\cF$ on $\cPC_1$ and $\cPC_2$ is a function that assigns any point $\vp\in \cPC_1$ and the transformed point $\vT(\vp)\in \cPC_2$ with the same value. A simple example for such an invariant feature is the one that assigns each point with its distance from the point cloud center of mass.
Since for finite support objects, it is straightforward to reduce the problem of computing the rigid transformation $\vT(\vp)=\vR\vp+\vt$ to a rotation-only problem, \ie, $\vt=0$, we next show that the moment integral calculations involved in evaluating the UME operator, may be replaced by computing moments of the invariant functions using summations:
\vspace{-2mm}
\begin{theorem}
Let $\vR$ be a rotation matrix and $\cPC_1$ and $\cPC_2$ be two point clouds satisfying the relation 
$ \cPC_2=\vR\cdot \cPC_1$.
Let $\cF$ be an $\SOG(3)$ invariant function on $\cPC_1$ and $\cPC_2$. Then,
\vspace{-2mm}
\begin{align}\label{eq:RrelationSUMS}
    \Mom_{\cPC_2}(\cF)=\vR \cdot  \Mom_{\cPC_1}(\cF),   \quad \text{where }\Mom_{\cPC_i}(\cF)=  \frac{1}{\abs{\cPC_i}} \begin{bmatrix}
    \sum_{\vp \in \cPC_i} p_1\cF(\vp)\\
    \sum_{\vp \in \cPC_i} p_2\cF(\vp)\\
    \sum_{\vp \in \cPC_i} p_3\cF(\vp)
    \end{bmatrix}, \
    \vp=
    \begin{bmatrix}
    p_1\\
    p_2\\
    p_3\\
    \end{bmatrix}
.
\end{align}
\end{theorem}
(See the supplementary material for the proof). We call $\Mom_{\cPC_i}$ the moment vector of the transformation invariant function  $\cF$ defined on $\cPC_i$.

Given two sets of points in $\R^3$,  $\mathset{\vv_i}_{i=1}^k$ and $\mathset{\vu_i}_{i=1}^k$ $k\geq 3$, satisfying the relation $\vv_i=\vR\cdot \vu_i$ for all $i$, we may find $\vR$ by a standard procedure proposed by Horn \etal \cite{horn1988closed}. Hence, we conclude from \eqref{eq:RrelationSUMS} that in the absence of noise, finding a set of invariant functions $\cF_1,\dots, \cF_k$, such that $k\geq 3$, yields a closed-form solution to the registration problem. 
However, in the presence of noise (sampling or additive) \eqref{eq:RrelationSUMS} no longer holds as $\cPC_2$ and $\vR \cdot \cPC_1$ are not identical anymore and in fact, with a high probability they do not share any point in common. Therefore, the estimated rotation matrix is noisy.  

This is the point where a deep neural network comes into play. 
The registration error obviously depends on the difference between $\Mom_{\cPC_2}$ and $\vR\cdot \Mom_{\cPC_1}$, and on  the  function $\cF$ being rigid transformation invariant despite the noise. To that extent, we employ a deep neural network in order to \textit{learn} how to construct good transformation-invariant functions. These invariant functions are designed to exploit the geometry of the point cloud so that their invariance to the transformation is minimally affected by the noise, resulting in a smaller registration error.

\textbf{Feature Extraction} 
Towards obtaining noise resilient $\SOG(3)$-invariant functions for effectively evaluating the UME moments, we aim at learning features capturing the geometric structure of the point cloud. This structure is determined by the point cloud coordinates (global information) and the neighborhood of each point (local information). 
We adopt the joint usage of DGCNN and Transformer blocks \cite{wang2019deep}, which has been proven to be very efficient in creating high-dimensional embedding  for point clouds, and adapt it to the UME framework in order to \textit{learn} multiple high-quality $\SOG(3) $ invariant features.


The DGCNN block is designed to preform a per-point embedding, such that information on neighboring points is well incorporated.
Each input point in the cloud is characterized by the coordinates of points in its neighborhood in addition to its own coordinates. This approach results in a new representation for each point using a $6 \times k$ matrix. Each column of that matrix represents one of the $k$ nearest neighbors and consists of the coordinates of the observed point stacked on top of the coordinated of the neighbor point.
DGCNN processes one point cloud at a time, sharing its weights between the two clouds. Ideally, our embedding network should result in identical feature (function value) for two corresponding points in the reference point cloud and in its transformed version. 

\textbf{SO(3)-invariant coordinate system followed by  joint-resampling} 
Since DGCNN weights are shared, when the coordinates of corresponding points between two point clouds are significantly different, the learning process fails. Clearly, this situation occurs when transformations of large magnitude, \eg large rotations, are considered (demonstrated in Section \ref{sec:Experiments}). Therefore DGCNN architecture in its original design, cannot be employed towards constructing invariant features when rotations by large angles are considered.

We overcome this inherent difficulty by mapping the input point clouds to an alternative coordinates system, which is $\SOG(3)$ invariant: Given a point cloud $\cPC$, the cloud center of mass (denoted by $\vm_{\cPC}$) is subtracted from each point coordinates, to obtain a centered representation $\cPC'$. We then construct a new coordinate system for the point cloud using PCA. That is, the axes of the new coordinate system are the principle vectors of the point cloud  covariance matrix given by 
\begin{equation}
\mathbf H^{\cPC'}=\sum_{\vp \in \cPC'}\vp\vp^T
\end{equation}
The principle vectors form the axes of the new coordinate system, and the new coordinates of each point are the projection coefficients on these axes. Formally, for a point $\vp\in \cPC'$, the new coordinates of $\vp$ are defined to be $ \vc_\vp=\vD^T \vp $ where $\vD$ is the matrix whose columns are the principle vectors. The resulting point cloud new coordinates are denoted by ${\cC}$.

It is easy to verify (see supplementary) that the new axes (columns of the PCA matrix) are co-variant under a rigid transformation. Therefore if $\cPC_2$ is obtained by a rigid transformation of $\cPC_1$, it holds that
$\cC_1=\cC_2$.
Furthermore, since the change of coordinate system is invertible, the original point cloud can be reconstructed.

However, in our setting, where the point clouds to be registered are sparse, differently sampled and noisy, the relation $\cC_1=\cC_2$ does not hold anymore. The loss of information caused by low sampling rate makes the resulting representations of the clouds significantly different yielding axes that are no longer co-variant and thus projections that are no longer invariant. 
Nonetheless, employing the $\SOG(3)$ invariant representation of the point clouds, the difference between $\cC_1$ and $\cC_2$ is sufficiently small to enable a learning process of multiple invariant features using a DGCNN block. 

While DCP strategy,\cite{wang2019deep}, is to jointly-resample the embedded point clouds in a high-dimensional feature space via a Transformer \cite{vaswani2017attention}. In DeepUME we  adopt the joint-resampling strategy, but DeepUME resamples the low-dimensional coordinate space, rather than  the high-dimensional feature space. 
We note that resampling point clouds in the coordinate space has a complexity advantage since the dimension of the coordinate space is 3, while the dimension of the embedding space is much higher  ($512$ in \cite{wang2019deep} and $32$ in our case).  In terms of architecture blocks, with this sampling approach the Transformer is leading the embedding network (and not the other way around, as in \cite{wang2019deep}).

We therefore employ the Transformer  for learning a resampling strategy of the projected samples in $\cC_1$ such that the resampling depends on the sampling of $\cC_2$, and vice versa.
Denoting the asymmetric function of the Transformer by $\phi$, we have
\begin{equation}
        \cC^\text{Transformer}_1=\cC_1 + \phi(\cC_1, \cC_2), \quad
        \cC^\text{Transformer}_2=\cC_2 + \phi(\cC_2, \cC_1).
\end{equation}
The additive terms, $\phi(\cC_1, \cC_2)$ and $\phi(\cC_2, \cC_1)$ are optimized to improve registration, by jointly "equalizing" the sampling patterns of both point clouds. 

The resampled point clouds produced using the Transformer are employed for evaluating the UME moments, by re-projecting them on the corresponding principle axes to obtain 
\begin{align}
    \widehat{\cPC}_1= \vD_1\cdot \cC^\text{Transformer}_1 +\vm_{\cPC_1}, \quad  
    \widehat{\cPC}_2= \vD_2\cdot \cC^\text{Transformer}_2 +\vm_{\cPC_2}. 
\end{align}
The point clouds, $\widehat{\cPC}_1$ and $\widehat{\cPC}_2$, are re-sampled versions of the original points clouds, related by the same rigid transformation that relates $\cPC_1$ and $\cPC_2$. We therefore apply the  UME registration to the resampled point clouds $\widehat{\cPC}_1$ and $\widehat{\cPC}_2$, using the  DGCNN generated functions designed to be $\SOG(3)$-invariant for differently and sparsely sampled  point clouds corrupted by noise. As demonstrated by the experimental results, applying UME registration on the re-projected re-sampled point clouds, provides with an improved performance. 

\textbf{Loss} In order to overcome the ambiguity problem in registering symmetric objects, discussed in Section \ref{sec:Experiments}, we adopt the Chamfer distance \cite{barrow1977parametric} as our loss function. That is, if $\hat{\vT}$ is the estimated transformation, 
\begin{align}
\mathcal{L}(\hat{\vT})=d_C(\hat{\vT}(\cPC_1),\cPC_2).
\end{align}
Using this loss,  ambiguous symmetric objects do not damage the learning process, as even if an ambiguity exists and the registration is successful, the Chamfer distance will be small. The Chamfer loss function has another advantage as no labels are required for the learning process, which makes it unsupervised.

\vspace{-4mm}
\section{Experiments}\label{sec:Experiments}
\vspace{-3mm}
We conduct experiments on three data sets: ModelNet40 \cite{wu20153d}, FAUST \cite{bogo2014faust} and Stanford 3D Scanning Repository \cite{StanfordScanRep} where the latter two are used only for testing. We train our network using ModelNet40, which consists of 12,311 CAD models in $40$ categories where $80\%$ samples are used for training and the rest for testing. 

Following previous works experimental settings, we uniformly sample 1,024 points from each model’s outer surface and further center and rescale the model into the unit sphere.
In our training and testing procedure, for each point cloud, we choose a rotation drawn uniformly from the full range of Euler angles and a translation vector in $[-0.5, 0.5]$ in each axis. We apply the rigid transformation obtained from the resulting parameters on $\cPC_1$, followed by a random shuffling of the points order, and get $\cPC_2$. In noisy scenarios, a suitable noise is applied to $\cPC_2$.
We train our framework in the scenario of Bernoulli noise (see bellow), in the specific case where $p_1=p_2=0.5$, and test all scenarios using the trained configuration.

Each experiment is evaluated using four metrics: The RMSE metric, both for the rotation and the translation
as well as the Chamfer distance and the Hausdorff distance \cite{liu2018point, DBLP:conf/eccv/UrbachBL20}, proposed next as alternative metrics to resolve ambiguity issues.
We compare our performances with the basic implementation of the UME method (coded by ourselves), ICP (implemented in Intel Open3D \cite{zhou2018open3d}), and four learned methods; PointNetLK and DCP (benchmarks point cloud registration networks) as well as the recently proposed DeepGMR, \cite{yuan2020deepgmr} and RGM, \cite{Fu2021RGM}. 
We retrain the baselines, adapting the code released by the authors. The experimental results are detailed below and summarized in Tables \ref{tbl:modelnet40_exp}, \ref{tbl:faust_stanford_exp}; further experimental results are provided in the supplementary.

\textbf{The ambiguity problem in the registration of symmetric objects} An ambiguity problem arises in point cloud registration whenever symmetric objects are considered, as more than a single rigid transformation (depending on the degree of symmetry of the object) can correctly align the two point clouds. 
In the noise free scenario such an ambiguity is trivially  handled since a fixed point constellation undergoes a rigid transformation which in the case of nonuniform sampling guarantees the uniqueness of the solution. However, when observations are noisy, the ambiguity is harder to resolve as the constellation structure breaks. In that scenario, $\cPC_2$ is a noisy version of $\vT(\cPC_1)$, and possibly no rigid transformation can perfectly align the point clouds. In that case, if the point clouds to be aligned represent a symmetric shape, there are multiple transformations that approximately align the two clouds together.

For symmetric shapes, which are very common in man-made objects, the rotation angles RMSE metric may assign large errors to successful registrations. To resolve this ambiguity we replace this metric by the Chamfer and Hausdorff distances defined by
\begin{align}\label{eq:chamferDef}
\begin{split}
    d_\text{C}(\cPC_1,\cPC_2)&=\frac{1}{\abs{\cPC_1}}\sum_{\vp\in \cPC_1} \min_{\vp'\in \cPC_2}\norm{\vp-\vp'}+\frac{1}{\abs{\cPC_2}}\sum_{\vp'\in \cPC_2} \min_{\vp\in \cPC_1}\norm{\vp'-\vp}\\
    d_\text{H}(\cPC_1,\cPC_2)&=\max_{\vp\in \cPC_1} \min_{\vp'\in \cPC_2}\norm{\vp-\vp'}+\max_{\vp'\in \cPC_2} \min_{\vp\in \cPC_1}\norm{\vp'-\vp}.
\end{split}
\end{align}

Using ModelNet40, we test performance in four scenarios: noise free model, sampling noise (Bernoulli noise and zero-intersection noise) and AWGN noise. We note that in the presence of sampling noise we indeed observe large errors in RMSE($\vR$) due to the symmetry of objects, although the symmetric shapes are well aligned. Moreover, in the unseen data sets, where  real world data is used, which is naturally asymmetric, the rotation RMSE is indeed small and indicates  successful registration. Therefore we consider the Chamfer and Hausdorff distances to be more reliable metrics for registration in the presence of symmetries. 

\textbf{ModelNet40: Noise free model}
We examine the case where no noise is applied to the measurements. In that case, we see that the estimation error is practically null. DeepGMR is shown to be the second best learned method, while all other tested methods yield large errors, meaning that the registration fails when large range of rotation angles is considered.     

\textbf{ModelNet40: Bernoulli noise}
The Bernoulli noise case is related to the scenario of  sampling noise, in which the point clouds contain different number of points. In that case, we choose randomly (uniformly) two numbers $p_1$ and $p_2$ in $[0.2, 1]$. Then, each point in $\cPC_i$ is removed with probability $1-p_i$, independently of the rest of the points. We perform registration on the resulting point clouds $\cPC_1^B$ and $\cPC_2^B$. We note that the number of points in $\cPC_i^B$ averages to $p_i\cdot 2048$, and is likely to be different in the two clouds. The number of corresponding points between the resulting clouds averages to $p_1p_2\cdot 2048$ (as the probability for the two point clouds to share a specific point is $p_1p_2$). We note that we were not able to evaluate RGM performance in that scenario.

\textbf{ModelNet40: Zero-intersection noise}
Zero intersection noise is the extreme (and most realistic) case of sampling noise. In that case, we randomly choose $1024$ points from $\cPC_1$ to be removed. Next, we remove the $1024$ points from $\cPC_2$ that correspond to the points  in $\cPC_1$. Thus, by construction, no point in one cloud is the result of applying a rigid transformation to a point in the other cloud. This scenario is visualized in Figure \ref{fig:teaser} that shows registration results of several baseline methods and our proposed method on a representative example from the unseen data set FAUST. 

\textbf{ModelNet40: AWGN}
\begin{figure*}
\centering
\includegraphics[width=\textwidth]{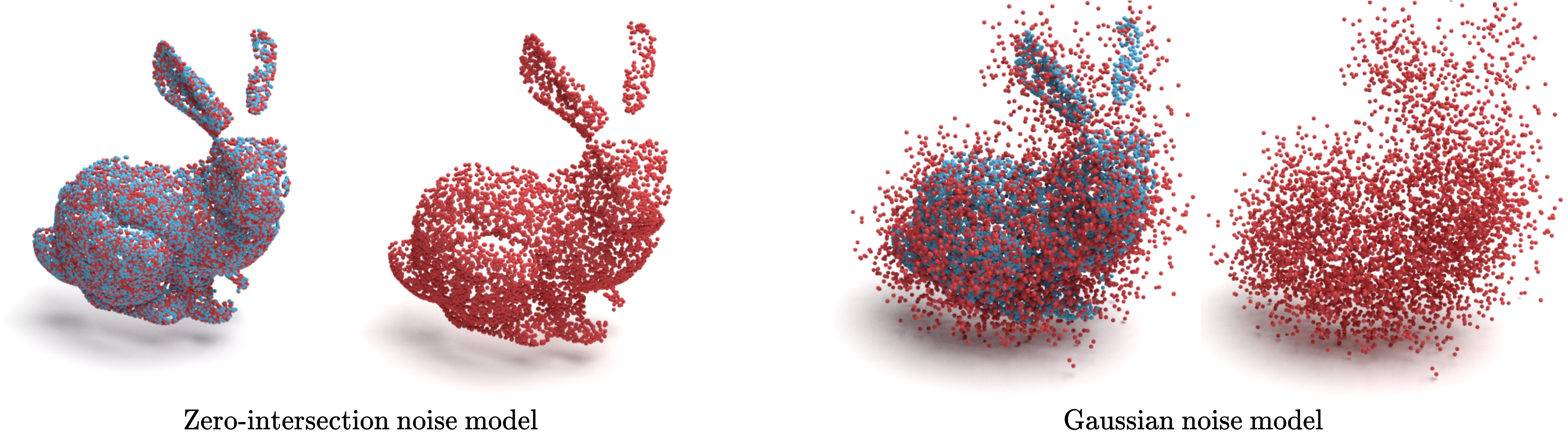}
\vspace{-3mm}
\caption{Sampling-noise vs. AWGN. Red point clouds are noisy observations of the blue ones, for zero-intersection (left) and AWGN with variance $0.09$ (right) noise models. While both achieve the same registration error under DeepUME, the former preserves well the original shape and the latter sever distortion makes it barely recognizable.}
\label{fig:gaussian_vs_zero_inter}
\end{figure*}
In these set of experiments each coordinate of every point in $\cPC_2$ is perturbed by an additive white Gaussian random variable drawn from  $\mathcal{N}(0,\sigma)$ where $\sigma$ is chosen randomly in $[0,0.04]$. All noise components are independent and no value clipping is performed.  The results in Table \ref{tbl:modelnet40_exp} indicate that since the UME is an integral (summation) operator the registration error of both the UME and the DeepUME is lower than the error of the other tested methods. 
We find that for sufficiently large additive noise variance, the rotation RMSE in both AWGN and sampling-noise models are comparable, yet these different types of noise have very different impact on registration, as illustrated in Figure \ref{fig:gaussian_vs_zero_inter}. In order to have the same registration error as in the zero-intersection model on the Stanford data set, an AWGN with variance of approximately 0.09 is required. As shown in Figure \ref{fig:gaussian_vs_zero_inter}, such a noise causes a sever distortion, which practically makes the original shape unrecognizable. On the other hand, the original shape of the bunny in the sampling noise scenario, is well preserved.

\begin{table}
\centering
\fontsize{7}{7}\selectfont
\setlength{\tabcolsep}{1.7pt} 
\renewcommand{\arraystretch}{1} 
\begin{tabular}{l cccc cccc cccc}
\midrule
& 
\multicolumn{4}{c}{Noise free} & 
\multicolumn{4}{c}{Bernoulli noise} &
\multicolumn{4}{c}{Gaussian noise}  \\

\cmidrule(lr){2-5} \cmidrule(lr){6-9} \cmidrule(lr){10-13}

\textbf{Model} & \textbf{$d_\text{C}$}  & \textbf{$d_\text{H}$}  & \textbf{RMSE}(R) & \textbf{RMSE}(t) 
& \textbf{$d_\text{C}$}  & \textbf{$d_\text{H}$} & \textbf{RMSE}(R) & \textbf{RMSE}(t)
& \textbf{$d_\text{C}$}  & \textbf{$d_\text{H}$} & \textbf{RMSE}(R) & \textbf{RMSE}(t)\\
\midrule

UME \cite{efraim2019universal}  
        & <1e-04    & <7e-04    & \underline{0.193}    & \underline{<1e-05}
        & 0.0581  	& 0.394 	& 74.164   & \underline{0.015}
        & 0.019	& 0.151    & \underline{27.684}    & \underline{0.002}  \\
        
ICP \cite{besl1992method}
        & 0.275 	& 1.446 	& 83.039	& 0.276
        & 0.297 	& 1.480	    & 86.381	& 0.286
        & 0.266	    &1.436  	& 83.073	& 0.277 \\
\midrule
PointNetLK \cite{aoki2019pointnetlk}  
        & 0.027 	& 0.142	    & 80.360	& 1.0105
        & \underline{0.029} 	& \underline{0.157} 	& 83.280    & 1.073
        & 0.026 	& 0.153 	& 81.843    & 1.037  \\
        
DCP \cite{wang2019deep}  
        & 0.059 	& 0.470	    & 92.285	& 0.014
        & 0.067 	& 0.483 	& 92.818	& 0.020
        & 0.055 	& 0.456 	& 90.715    & 0.014 \\
        
DeepGMR \cite{yuan2020deepgmr}  
        & \underline{<7e-06}	& \underline{<9e-05}  	& \underline{0.193} 	& <5e-05
        & 0.033 	& 0.224 	& \underline{72.447}	& 0.018
        & \underline{0.011} 	& \underline{0.085}	    & 42.515	& 0.004 \\
        
RGM \cite{Fu2021RGM}  
        & 0.255   & 1.333   & 99.937   & 0.388
        & \text{N/A}  & \text{N/A}   & \text{N/A}   & \text{N/A}
        & 0.254   & 1.331   & 100.117   & 0.389  \\
       
\midrule
DeepUME (ours) 
        & \textbf{<1e-07}    & \textbf{<1e-07}    & \textbf{<3e-04}    & \textbf{<1e-07}
        & \textbf{0.010}     & \textbf{0.083} 	& \textbf{40.357}    & \textbf{0.015 }	
        & \textbf{0.002} 	& \textbf{0.012}	    & \textbf{2.425}      & \textbf{0.001}\\

\midrule
\end{tabular}
\caption{ModelNet40 experimental results. Our method achieves substantial performance gains compared to the competing techniques for all metrics in all the examined scenarios. \textbf{$d_\text{C}$} and \textbf{$d_\text{H}$} stand for Chamfer and Hausdorff distances respectively. The best results are \textbf{bold} and the second best are \underline{underlined}.}
\label{tbl:modelnet40_exp}
\vspace{1mm}
\end{table}

\textbf{Unseen data sets}
The generalization ability of a model to unseen data is an important aspect for any learning based framework. In order to demonstrate that our framework indeed generalizes well, we test it on two unseen datasets. The first is the FAUST data set that contains human scans of 10  different subjects in 30 different poses each with about 80,000 points per shape, and the other is the Stanford 3D Scanning Repository. 
We generate the objects to be registered using a similar methodology to that employed for the ModelNet40 data set.
Our framework achieves accurate registration results in all scenarios checked, and shows superior performance over the compared methods. 

\begin{table}
\centering
\fontsize{7}{7}\selectfont
\setlength{\tabcolsep}{1.7pt} 
\renewcommand{\arraystretch}{1} 
\begin{tabular}{l cccc cccc cccc}
\midrule
& 
\multicolumn{4}{c}{ModelNet40 \cite{wu20153d}} & 
\multicolumn{4}{c}{FAUST \cite{bogo2014faust}} &
\multicolumn{4}{c}{Stanford 3D Scanning Repository \cite{StanfordScanRep}}  \\

\cmidrule(lr){2-5} \cmidrule(lr){6-9} \cmidrule(lr){10-13}

\textbf{Model} & \textbf{$d_\text{C}$}  & \textbf{$d_\text{H}$}  & \textbf{RMSE}(R) & \textbf{RMSE}(t) 
& \textbf{$d_\text{C}$}  & \textbf{$d_\text{H}$} & \textbf{RMSE}(R) & \textbf{RMSE}(t)
& \textbf{$d_\text{C}$}  & \textbf{$d_\text{H}$} & \textbf{RMSE}(R) & \textbf{RMSE}(t)\\
\midrule

UME \cite{efraim2019universal}  
        & 0.051 	& 0.373 	& 80.331	& \underline{0.010}
        & 0.007 	& 0.085 	& 35.983	& 0.044
        & 0.033 	& 0.267 	& 48.716	& \underline{0.010} \\
        
ICP \cite{besl1992method}
        & 0.276 	& 1.448 	& 82.948	& 0.277
        & 0.376	    & 1.643 	& 84.544	& 0.279
        & 0.288 	& 1.337 	& 87.292    & 0.277 \\
\midrule
PointNetLK \cite{aoki2019pointnetlk}  
        & 0.028 	& 0.147 	& 80.858	& 1.023
        & 0.018 	& 0.170	    & 90.512	& 1.120
        & 0.040 	& 0.289 	& 84.520	& 1.147
  \\
        
DCP \cite{wang2019deep}  
        & 0.059 	& 0.475 	& 93.221	& 0.014
        & 0.046 	& 0.516 	& 94.315	& 0.137	
        & 0.072 	& 0.522 	& 99.328	& 0.011 \\
        
DeepGMR \cite{yuan2020deepgmr}  
        & \underline{0.026}	    & \underline{0.117} 	& \textbf{67.282}	& \underline{0.010}
        & \underline{0.003}   	& \underline{0.027} 	& \underline{27.941}	& \underline{0.020}
        & \underline{0.005} 	& \underline{0.119} 	& \underline{39.402}	& 0.012\\
        
RGM \cite{Fu2021RGM}  
        & 0.254   & 1.335   & 100.970   &0.388
        & 0.385   & 1.677   & 114.496   &0.418 
        & 0.278   & 1.257   & 104.872   &0.368   \\
\midrule
DeepUME (ours) 
        & \textbf{0.011} 	& \textbf{0.094} 	& \underline{70.818}	& \textbf{0.009}
        & \textbf{0.002} 	& \textbf{0.024}	& \textbf{8.630}	& \textbf{0.019}
        & \textbf{0.002} 	& \textbf{0.110} 	& \textbf{5.625} 	& \textbf{0.010}\\
\midrule
\end{tabular}
\caption{Zero-intersection noise results on seen (ModelNet40) and unseen data sets (FAUST and Stanford 3D Scanning Repository). Our method outperforms the competing techniques in all scenarios for all metrics, only except RMSE($\vR$) for ModelNet40.}
\label{tbl:faust_stanford_exp}
\vspace{-1mm}
\end{table}

\vspace{-4mm}
\section{Conclusions}
\vspace{-3mm}
We derived a novel solution to the highly practical problem of aligning sparsely and differently sampled point clouds in the presence of noise and large transformations. 
Our model, named DeepUME, integrates the closed-form UME registration method into a DNN framework.
The two are combined into a single unified framework, trained end-to-end and in an unsupervised manner.
DeepUME employs an $\SOG(3)$-invariant coordinate system to learn both a joint-resampling strategy of the point clouds and $\SOG(3)$-invariant features. 
The resampling is performed in the low-dimensional coordinate space (rather than in the high-dimensional feature space), to minimize the effect of the sampling noise on the registration performance. 
The constructed features are utilized by the geometric UME method for transformation estimation.
The parameters of DeepUME are optimized using a metric designed to overcome the ambiguity emerging  in the registration of symmetric shapes, when noisy scenarios are considered. We show that our hybrid method outperforms state-of-the-art registration methods in various scenarios, and generalizes well to unseen datasets. Future work will extend the proposed method for registration of sub-parts and key-points, incorporating both local and global information to allow the registration of partially overlapping scenes.

\bibliography{ms}
\end{document}


\maketitle
\section{Overview}
This document  contains the following:
\begin{itemize}
    \item Section \ref{sec:proof}: Mathematical proofs of the statements mentioned in the main paper
    \item Section \ref{sec:added_rslts}: Additional registration qualitative and quantitative results
    \item Section \ref{sec:ablations}: Several ablation experiments are presented, replacing components of DeepUME by alternatives in order to evaluate the contribution of the proposed construction
    \item Section \ref{sec:exp_sting}: DeepUME architecture details
\end{itemize}


\section{Mathematical Proofs}\label{sec:proof}

\subsection{Discrete UME}
Relying on the original UME method, we prove the closed form formula presented in the paper in equation (1). 

\begin{theorem}
\label{thm:DUME}
Let $\vR$ be a rotation matrix and $\cPC_1$ and $\cPC_2$ be two point clouds satisfying the relation 
$ \cPC_2=\vR\cdot \cPC_1$.
Let $\cF$ be an $\SOG(3)$ invariant feature on $\cPC_1$ and $\cPC_2$, namely 
\begin{align}
\cF(\vp)=\cF(\vR \vp), \quad \forall \vp\in \cPC_1,
\end{align}
then 
\begin{align}
\label{eq:RrelationSUMS}
    \Mom_{\cPC_2}(\cF)=\vR \cdot  \Mom_{\cPC_1}(\cF),   \quad \text{where }\Mom_{\cPC_i}(\cF)=  \frac{1}{\abs{\cPC_i}} \begin{bmatrix}
    \sum_{\vp \in \cPC_i} p_1\cF(\vp)\\
    \sum_{\vp \in \cPC_i} p_2\cF(\vp)\\
    \sum_{\vp \in \cPC_i} p_3\cF(\vp)
    \end{bmatrix}, \
    \vp=
    \begin{bmatrix}
    p_1\\
    p_2\\
    p_3\\
    \end{bmatrix}
.
\end{align}
\end{theorem}

In order to prove Theorem \ref{thm:DUME}, we first state and prove a discrete version of the UME theorem from which Theorem \ref{thm:DUME} follows immediately as a special case. We begin by presenting the continuous UME theorem, \cite{efraim2019universal}, in the specific case where the translation vector $\vt=0$.

\begin{definition}
Let $k:\R^n\to \R$ be a compactly supported measurable function and $w_1,\dots,w_D: \R\xrightarrow{} \R$ are measurable functions.  The $n\times D$ UME matrix of $k$ with respect to $w_1,\dots,w_D$ is defined by
\begin{align}
  [\UME_k]_{i,j}:= \int_{\R^n}x_i w_j(k(\vx))d\vx. 
\end{align}
\end{definition}
 
\begin{theorem} \textbf{(Continuous UME)}
\label{thm:CUME}
\cite{efraim2019universal}, 
Let $f,g:\R^n\to \R$, be two functions with compact supports related by a rotation $\vR$, i.e. $g(\vx) = f(\vR\vx)$ for all $\vx\in \R^n$. Then, for any set of $D$  measurable functions $w_1,\dots,w_D$ such that $w_i(0)=0$ for all $i$,
\begin{align}
    \UME_f = \vR\cdot\UME_g.
\end{align}
\end{theorem}

We next provide the discrete analog of Theorem \ref{thm:CUME} where the continuous  functions $f$ and $g$ are replaced by the invariant functions estimated from the observed point clouds using the DNN, and integration is replaced by summation on the elements in the point clouds.

\begin{definition}
Let $\cPC\subseteq \R^3$ be a finite point cloud, $\cF$ a feature (function) on $\cPC$ and $w_1,\dots,w_d:\R\to \R$ measurable functions. The discrete $n\times D$ UME matrix of $\cPC$ and $\cF$ with respect to $w_1\dots,w_D$ is defined to be  
\begin{align}
    [\UME_{\cPC}^\cF]_{i,j}= 
    \sum_{\vp \in \cPC} p_i w_j(\cF(\vp)).
\end{align}
\end{definition}

\begin{proposition}
\label{prop:dUME}
Let  $\cPC_1$ and $\cPC_2$ be two point clouds satisfying $\cPC_2=\vR\cPC_1$, and $\cF$ is an invariant feature on $\cPC_1$ and $\cPC_2$. For any $D$ functions $w_1,\dots,w_d:\R\to\R$ satisfying $w_i(0)=0$ for all $i$ 
\begin{align}
\label{eq:discUMErelation}
    \UME_{\cPC_2}^\cF=\vR \cdot \UME_{\cPC_1}^\cF.
\end{align}
\end{proposition}

We note that if we take $w_1$ to be the identity function and denote the first column of $\UME_{\cPC_1}^\cF$ and $\UME_{\cPC_2}^\cF$ by $[\UME_{\cPC_1}^\cF]_1$ and $[\UME_{\cPC_2}^\cF]_1$ respectively, we have 
\begin{align}
 \frac{1}{\abs{\cPC_i}}[\UME_{\cPC_i}^\cF]_1=\Mom_{\cPC_i}(\cF), \quad i=1,2. 
 \end{align}
Hence, Theorem \ref{thm:DUME} is followed by Proposition \ref{prop:dUME} trivially, as a special case.

\textbf{Proof of Proposition \ref{prop:dUME}} The main idea in our proof is to approximate the discrete sums defining the discrete UME matrices in  \eqref{eq:discUMErelation} by continuous integrals and apply the continuous UME theorem. Given $\varepsilon>0$, $\cPC_1$, $\cPC_2$ and the invariant feature $\cF$, we construct two compactly supported measurable functions $f_{\varepsilon},g_{\varepsilon}:\R\to \R$ such that $g_{\varepsilon}(\vx)=f_{\varepsilon}(R^{-1}\vx)$. We then apply Theorem \ref{thm:CUME} to $f_\varepsilon$ and $g_\varepsilon$, and taking $\varepsilon\to 0$ we will conclude.

\begin{figure}
  \centering
  \includegraphics[width=0.7\textwidth]{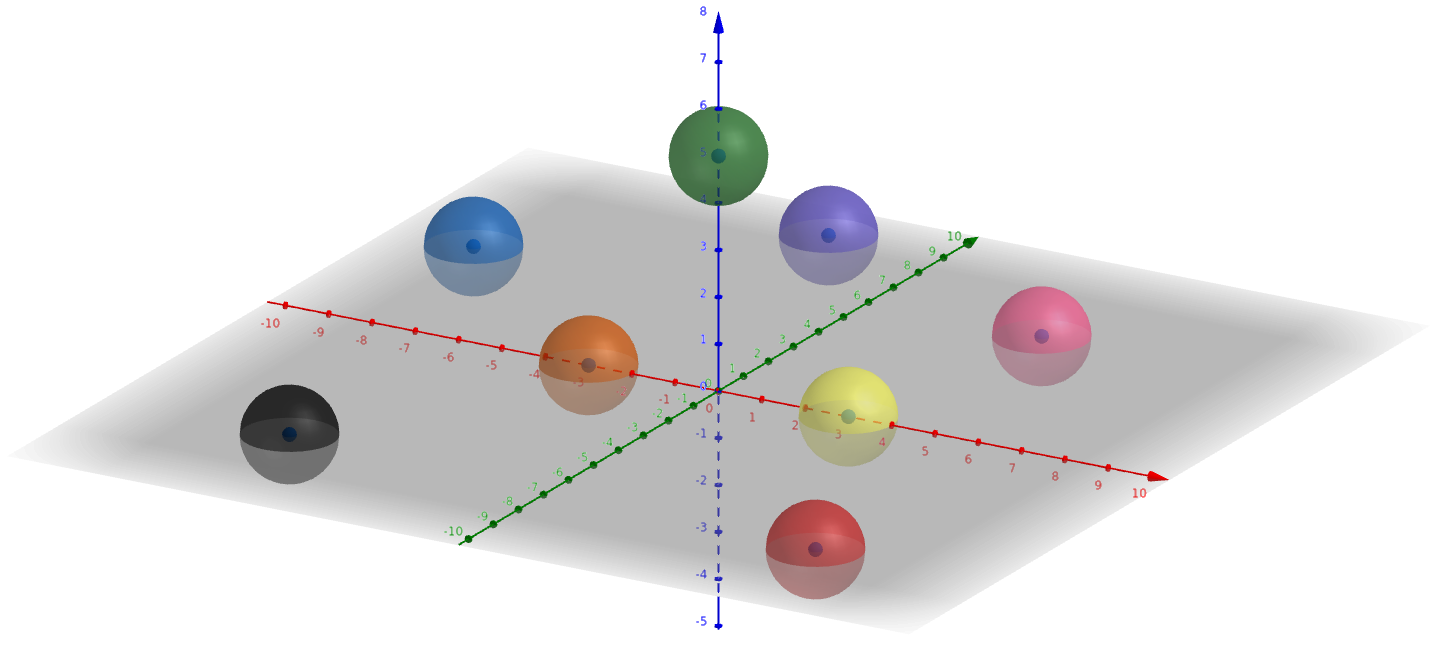}
  \caption{$B_{\varepsilon(\vp)}$ for $\vp \in \cPC$. Balls centers are the point cloud points, and each ball color represents the value of $\cF(\vp)$.}
  \label{fig:balls}
\end{figure}

Denote the ball of radius $\varepsilon$ centered at a point $\vp \in \R^3$ by $B_{\varepsilon}(\vp)$ . Define the functions $f_{\varepsilon}$ is and $g_{\varepsilon}$ by
\begin{align}\label{eq:ball_func}
    f_{\varepsilon}(\vx):=\sum_{\vp\in\cPC_1} \cF(\vp)\indicator_{B_{\varepsilon}(\vp)}(\vx), \quad g_{\varepsilon}(\vx):=\sum_{\vp\in\cPC_2} \cF(\vp)\indicator_{B_{\varepsilon}(\vp)}(\vx).
\end{align} See the illustration of $f_{\varepsilon}(\vx)$ in Figure \ref{fig:balls}. We now prove that the desired relation $g_{\varepsilon}(\vx)=f_{\varepsilon}(\vR^{-1}\vx)$ holds: Directly from the definition of $f_{\varepsilon}$,
\begin{align}\label{eq:disc_cont}
    f_{\varepsilon}(\vR^{-1}\vx)\overset{\eqref{eq:ball_func}}{=}
    \sum_{\vp\in\cPC_1} \cF(\vp)\indicator_{B_{\varepsilon}(\vp)}(\vR^{-1}\vx).
\end{align} Since a rigid transformation maps any ball to a ball with the same radius, we have that   $\indicator_{B_{\varepsilon}(\vp)}(\vR^{-1}\vx)$ is non-zero if and only if $\vR^{-1}\vx \in B_{\varepsilon}(\vp)$:
\begin{align}
    \vR^{-1}\vx \in B_{\varepsilon}(\vp) \iff
    \vx \in \vR(B_{\varepsilon}(\vp)) \iff
    \vx \in B_{\varepsilon}(\vR\vp).
\end{align}
It is immediately follows that 
\begin{align}
    \label{eq:point_in_ball}
    \indicator_{B_{\varepsilon}(\vp)}(\vR^{-1}\vx)= \indicator_{B_{\varepsilon}(\vR\vp)}(\vx), \quad \forall \vx\in \R^3.
\end{align}
Substituting \eqref{eq:point_in_ball} into \eqref{eq:disc_cont}, and using the $\SOG(3)$ invariance of $\cF$ we have
\begin{align}\label{eq:fp1_and_fp2}
    f_{\varepsilon}(\vR^{-1}\vx) &=
    \sum_{\vp\in\cPC_1} \cF(\vp)\indicator_{B_{\varepsilon}(\vp)}(\vR^{-1}\vx)\\
    &=
    \sum_{\vp\in\cPC_1} \cF(\vR\vp)\indicator_{B_{\varepsilon}(\vR\vp)}(\vx)\\
    &=
    \sum_{\vp\in\cPC_2} \cF(\vp)\indicator_{B_{\varepsilon}(\vp)}(\vx)=
    g_{\varepsilon}(\vx).
\end{align}
We have so far proved that $f_{\varepsilon}(\vx)=g_{\varepsilon}(\vR\vx)$ for all $\vx$. By Theorem \ref{thm:CUME} we have
\begin{align}\label{eq:ume_fpc}
\UME_{g_{\varepsilon}} = \vR\cdot\UME_{f_{\varepsilon}}.
\end{align}

For sufficiently small $\varepsilon$, the balls defining $f_{\varepsilon}$ do not intersect and therefore we have,
\begin{align}
    [\UME_{f_{\varepsilon}}]_{ij}&\overset{\eqref{eq:ball_func}}{=}\int_{\R^3}x_i w_j
    \parenv*{
    \sum_{\vp\in\cPC_1} \cF(\vp)\indicator_{B_{\varepsilon}(\vp)}(\vx)}d\vx\\
    &=\int_{\biguplus_{\vv\in\cPC_1} B_{\varepsilon}(\vv)} x_i w_j\parenv*{
    \sum_{\vp\in\cPC_1} \cF(\vp)\indicator_{B_{\varepsilon}(\vp)}(\vx)}d\vx\\
    & \quad +\int_{\R^3 \backslash \biguplus_{\vv\in\cPC_1} B_{\varepsilon}(\vv)}x_i \underbrace{w_j\parenv*{
    \sum_{\vp\in\cPC_1} \cF(\vp)\underbrace{\indicator_{B_{\varepsilon}(\vp)}(\vx)}_{=0}}}_{w_j(0)=0}d\vx  \\
    &=\sum_{\vv\in\cPC_1}\int_{B_\varepsilon(\vv)} x_i w_j\parenv*{
    \sum_{\vp\in\cPC_1} \cF(\vp)\indicator_{B_{\varepsilon}(\vp)}(\vx)}d\vx
    \\
    &=\sum_{\vv\in\cPC_1} w_j(\cF(\vv))
    \int_{B_\varepsilon(\vv)} x_id\vx.\label{eq:disc_ume_entry}
    \end{align} 
where $\biguplus $ denotes a disjoint union of sets and the last equality stems from the fact that  $\cF(\vp)$ is constant on ${B_\varepsilon(\vv)}$. By \eqref{eq:disc_ume_entry} and the integral mean value theorem we have that \begin{align}\label{eq:int_to_sum}
   \begin{split}
    \lim_{\varepsilon \to 0} \frac{1}{\text{Vol}(B_\varepsilon)}[\UME_{f_{\varepsilon}}]_{ij}
    &=\lim_{\varepsilon \to 0} \frac{1}{\text{Vol}(B_\varepsilon)}
    \sum_{\vp\in\cPC_1} w_j(\cF(\vp))
    \int_{B_\varepsilon(\vp)} x_id\vx\\
    &=\sum_{\vp\in\cPC_1} w_j(\cF(\vp))
    \underbrace{\lim_{\varepsilon \to 0} \frac{1}{\text{Vol}(B_\varepsilon)}\int_{B_\varepsilon(\vp)} x_id\vx}_{p_i}
    =\sum_{\vp\in\cPC_1} p_i w_j(\cF(\vp))
    \end{split}
\end{align} This shows that 
\begin{align}
    \UME_{\cPC_1}^\cF=\lim_{\varepsilon \to 0} \frac{1}{\text{Vol}(B_\varepsilon)}\UME_{f_{\varepsilon}},
\end{align}
and similarly it is easily proved that 
\begin{align}
    \UME_{\cPC_2}^\cF=\lim_{\varepsilon \to 0} \frac{1}{\text{Vol}(B_\varepsilon)}\UME_{g_{\varepsilon}}.
\end{align}

Finally, applying Theorem \ref{thm:CUME} on $f_{\varepsilon}$ and $g_{\varepsilon}$ we obtain: 
\begin{align}
   \UME_{\cPC_2}^{\cF}&=     
   \lim_{\varepsilon \to 0} \frac{1}{\text{Vol}(B_\varepsilon)}\UME_{g_{\varepsilon}} = 
    \lim_{\varepsilon \to 0} \frac{1}{\text{Vol}(B_\varepsilon)}
    \vR\cdot \UME_{f_{\varepsilon}}\\
    &=\vR\cdot  \lim_{\varepsilon \to 0} \frac{1}{\text{Vol}(B_\varepsilon)}
    \UME_{f_{\varepsilon}}=R\cdot \UME_{\cPC_1}^{\cF}.
\end{align}
This completes the proof.

\subsection{$\SOG(3)$-invariant coordinate system}\label{sec:PCA}
In the proposed method, point cloud raw coordinates are mapped to a transformation invariant representation by projecting them on the coordinate system defined by the principle vectors of the point cloud PCA. 

More specifically, given a point cloud $\cPC$, the cloud center of mass (denoted by $\vm_{\cPC}$) is subtracted from each point coordinates, to obtain a centered representation $\cPC'$. The axes of the new coordinate system are the principle vectors of the point cloud  covariance matrix given by 
\begin{equation}
\vH_{\cPC'}=\sum_{\vp \in \cPC'}\vp\vp^T
\end{equation}
For a point $\vp\in \cPC'$, the new coordinates of $\vp$ are defined to be $\vc_\vp=\vD_{\cPC'}^T \cdot \vp $ where $\vD_{\cPC'}$ is a matrix whose columns are principle vectors. Formally, $\vD_{\cPC'}$ is an orthogonal matrix for which $\vH_{\cPC'}=\vD_{\cPC'}{\bm\Lambda} \vD_{\cPC'}^T$ for a diagonal matrix ${\bm\Lambda}$. The resulting point cloud new coordinates are denoted by ${\cC}$.

We shall now verify that the new axes (columns of the PCA matrix) are  rotation co-variant:
\begin{align}
    \vH_{\vR\cPC'}=\sum_{\vp \in \vR\cPC'} \vR \vp \vp^T \vR^T=
    \vR\vH_{\cPC'}\vR^T=\vR\vD_{\cPC'}{\bm\Lambda}(\vR\vD_{\cPC'})^T.
\end{align}
That is, 
\begin{align}
\label{eq:axes_covariant}
    \vD^{\vR\cPC'}=\vR\vD_{\cPC'}.
\end{align}
Using \eqref{eq:axes_covariant}, we easily prove that the projection coefficients on the new axes are rotation invariant. Given $\cPC_1$ and $\cPC_2$ related by a rigid motion, we have
\begin{align}\label{eq:clouds_invariance}
    \begin{split}
    \cC_1&=\bracenv*{\parenv*{\vD_{\cPC'_1}}^T\vp:\vp\in \cPC_1}
    =\bracenv*{\parenv*{\vR^T\vD_{\cPC'_2}}^T\vp:\vp\in \cPC_1}\\
    &
    =\bracenv*{\parenv*{\vD_{\cPC'_2}}^T\vR\vp:\vp\in \cPC_1}=\bracenv*{\parenv*{\vD_{\cPC'_2}}^T\vp:\vp\in \cPC_2}
    =\cC_2.
    \end{split}
\end{align}

\subsubsection{Axes sign ambiguity}
For a point cloud $\cPC$,  the principal vectors defining $\vD_{\cPC'}$, are defined up to a sign. Hence,  the equality in \eqref{eq:axes_covariant} is true up to multiplication of the the columns by $\pm 1$. That is to say, only one from the 8 possibilities for the principal vectors matrix satisfies the desired equality \eqref{eq:axes_covariant}.
We eliminate this sign ambiguity by considering all $8$ possible new axes systems ${\bracenv*{D_{\cPC'_2}^i}}_{i=1}^8$ given by different sign multiplication constellations. We choose the one axes system that satisfies  \eqref{eq:clouds_invariance} by 
\begin{align}
i = \argmin_{j=1,\dots,8} d_{\text{C}}(\cC_1, \cC^j_2) \end{align}
where $d_{\text{C}}$ stands for Chamfer distance.


\section{Additional Registration Results}\label{sec:added_rslts}

\subsection{The Effect of Sampling-Rate on Registration under Sampling-Noise Scenarios}
\begin{figure}
    \centering
    \begin{minipage}{0.45\linewidth}
    \includegraphics[width=\textwidth]{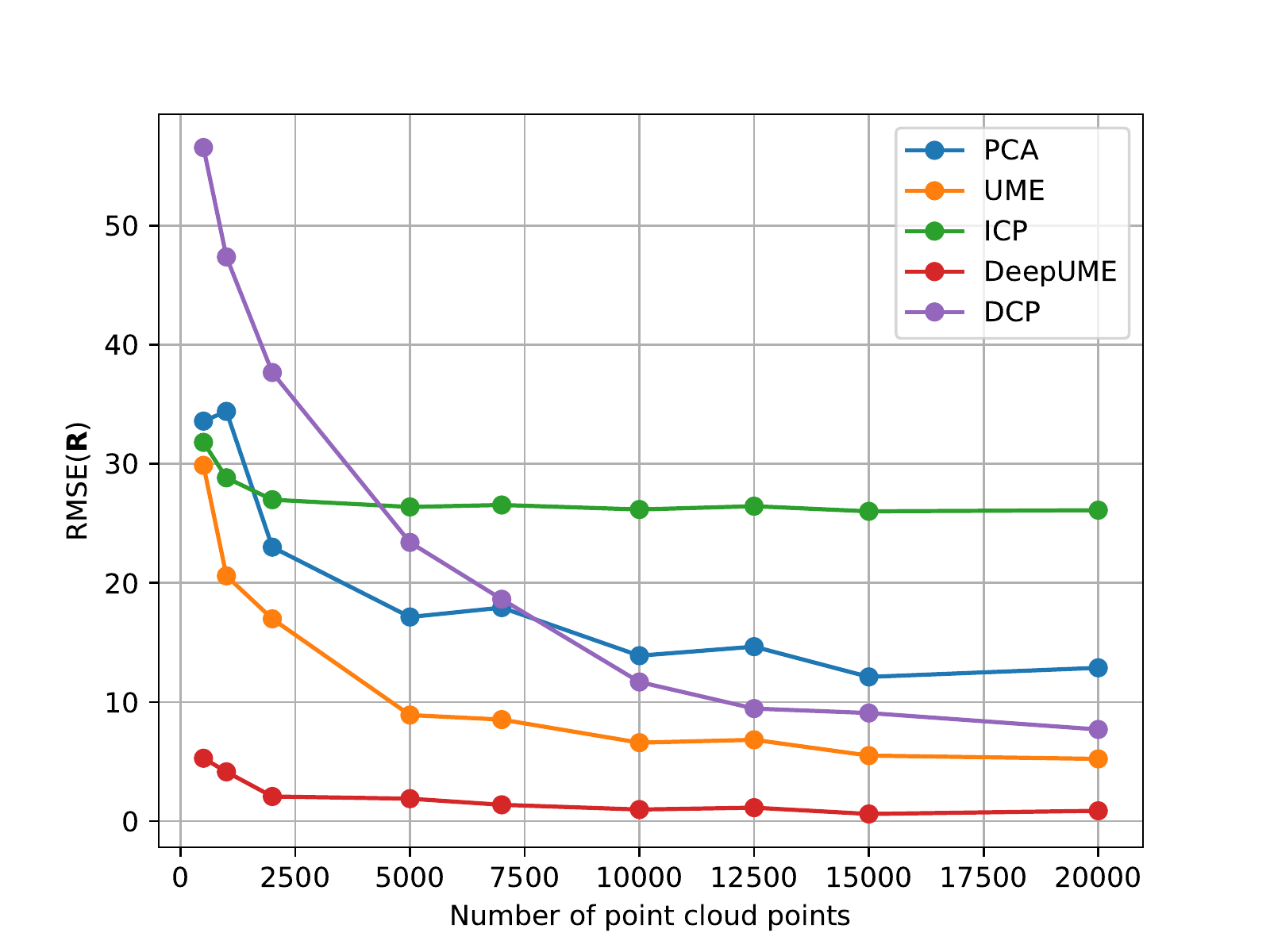}
    \caption{Registration performance under zero-intersection noise model and different point densities on the FAUST data set. On dense point clouds, the compared registration methods achieve comparable results.  However,  as the sampling rate decreases, the sampling noise effect becomes dominant and the performance of all methods, but DeepUME, severely deteriorates.}
    \label{fig:dense_sparse}
    \end{minipage}\quad
    \begin{minipage}{0.45\linewidth}
    \includegraphics[width=1\textwidth]{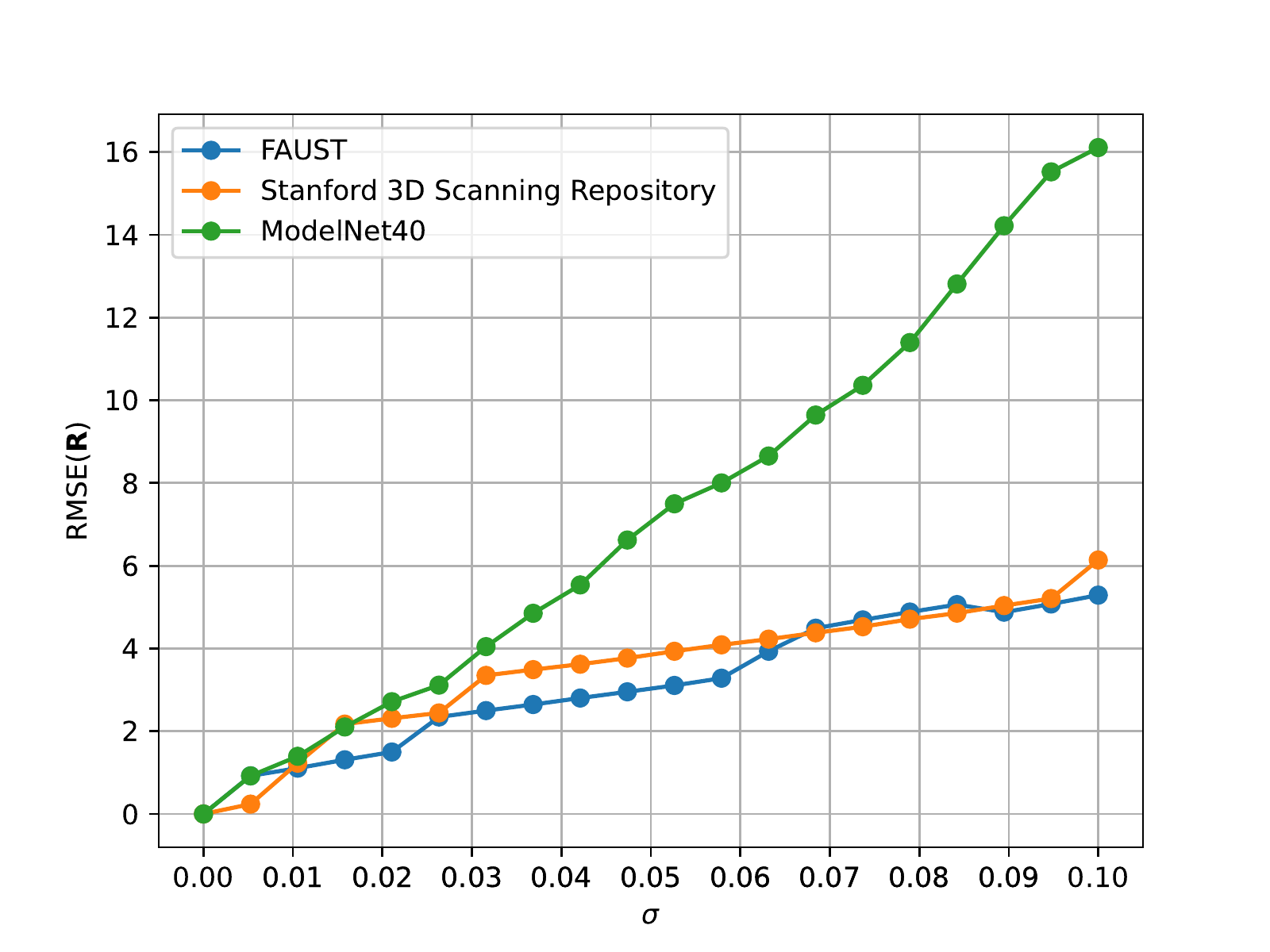}
    \caption{The effect of the WAGN variance on DeepUME performance (measured in RMSE($\vR$)) in all three data sets tested.
    In the presence of noise, more symmetric examples in ModelNet40 data set are incorrectly registered and therefore the average error increases dramatically.}
    \label{fig:gaussian}
    \end{minipage}
\end{figure}

{
Figure \ref{fig:dense_sparse} depicts the registration performance of different methods under different point densities. 

As mentioned in the introduction of the main paper, many point cloud registration applications process under-sampled point clouds. 
Figure \ref{fig:dense_sparse} shows that on dense point clouds, the evaluated registration methods achieve comparable results. However,  as the sampling rate decreases, the sampling noise effect becomes dominant and the performance of all methods, but DeepUME, severely deteriorates.

We note that for the closed form registration methods considered, where computational requirements are moderate, registration is testable with up to 80,000 sample points per cloud.  In that scenario, PCA and UME registration achieve RMSE$(\vR)$ errors of 5.147 and 2.573 respectively. This is particularly interesting since it shows that PCA and UME, which are both critical parts in our framework, require about \textit{160 times more samples than DeepUME} (80k vs 500) in order to achieve similar performances.
It is further implied that the main effect of the proposed method is equivalent to the interpolation of sub-sampled point clouds where the quality criterion of the interpolation is its efficiency for registration purposes.}

\subsection{Symmetric objects registration ambiguity}\label{sec:ambiguity}
\begin{figure}
  \centering
  \includegraphics[width=0.9\textwidth]{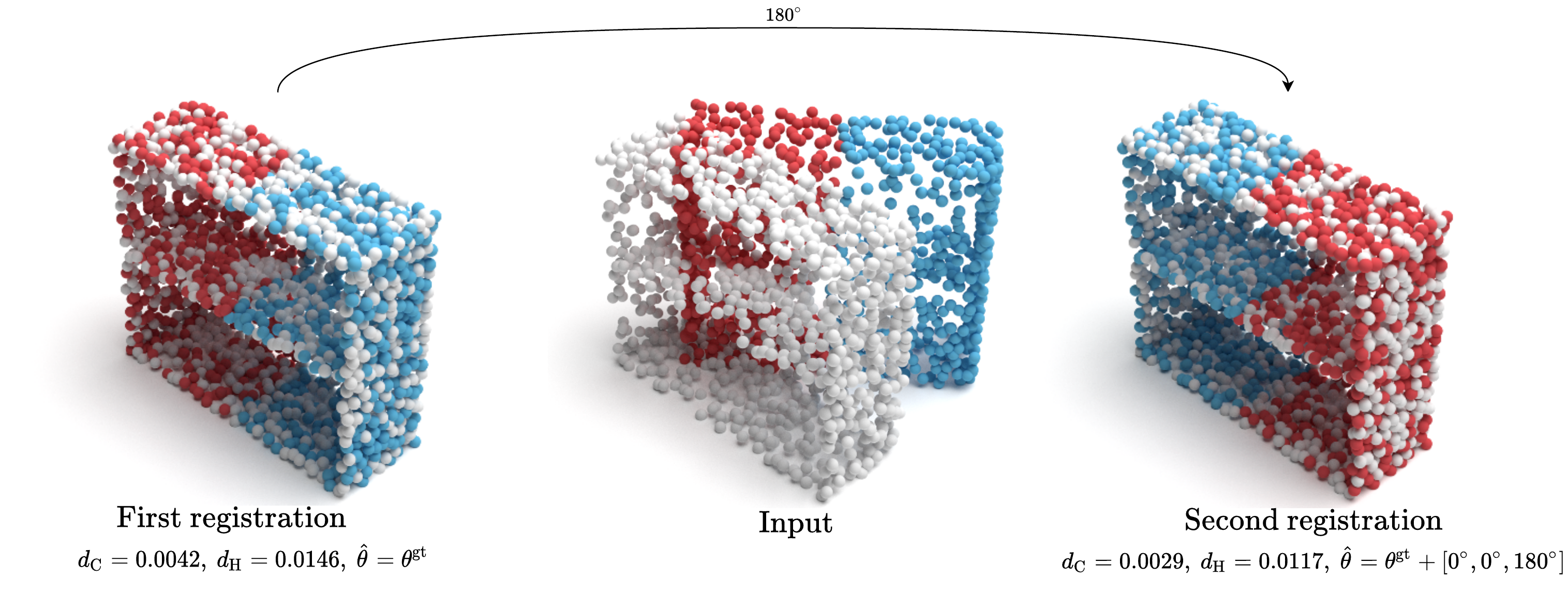}
  \caption{Symmetric objects registration ambiguity arising in sampling noise scenarios. Due to the sampling noise there is no one true registration solution for the input, but any that aligns the clouds together. Therefore, the two presented solutions should result in a low error metric, which is reflected in the Chamfer and Hausdorff distances and not in RMSE($\vR$).}
  \label{fig:ambiguity}
\end{figure}

In Figure \ref{fig:ambiguity}, we demonstrate the ambiguity problem discussed in section 5 in the paper. The box shaped bookshelf (as many other examples in ModelNet40) is symmetric under rotations of $180^\circ$ about the $z$ axes. Therefore, as shown the in Figure \ref{fig:ambiguity}, in the sampling noise scenario (zero-intersection model) two possible registration solutions (that differ by a $180^\circ$ rotation about the $z$ axis) are possible. Both solutions do align the point clouds, yet one achieves zero RMSE($\vR$) error, while the other yields a much higher one. Nevertheless, both 
Chamfer and Hausdorff distances achieve small errors which implies they are better suited  for measuring  registration error on symmetric shapes. 

\subsection{Gaussian Noise}
One of the most intriguing observations presented in the experimental section considers the effect of different types of noise on registration, and in particular, Additive White Gaussian Noise (AWGN) versus sampling noise. In Figure, \ref{fig:gaussian} we evaluate the effect of the noise variance on the rotation RMSE. An interesting observation is that registration on ModelNet40 data set \cite{wu20153d} is significantly more affected by AWGN, than other types of data. Recalling the ambiguity problem of ModelNet40 (see section \ref{sec:ambiguity} and section 5 in the main paper) this is quite expected: Noise makes the inherent ambiguity problem harder to resolve. In the presence of noise, many of the  ambiguous  examples are falsely registered and therefore the average error increases dramatically.

\subsection{Bernoulli Noise}
\begin{figure}
    \centering
    \begin{minipage}{\linewidth}
    \centering
    \includegraphics[width=\textwidth]{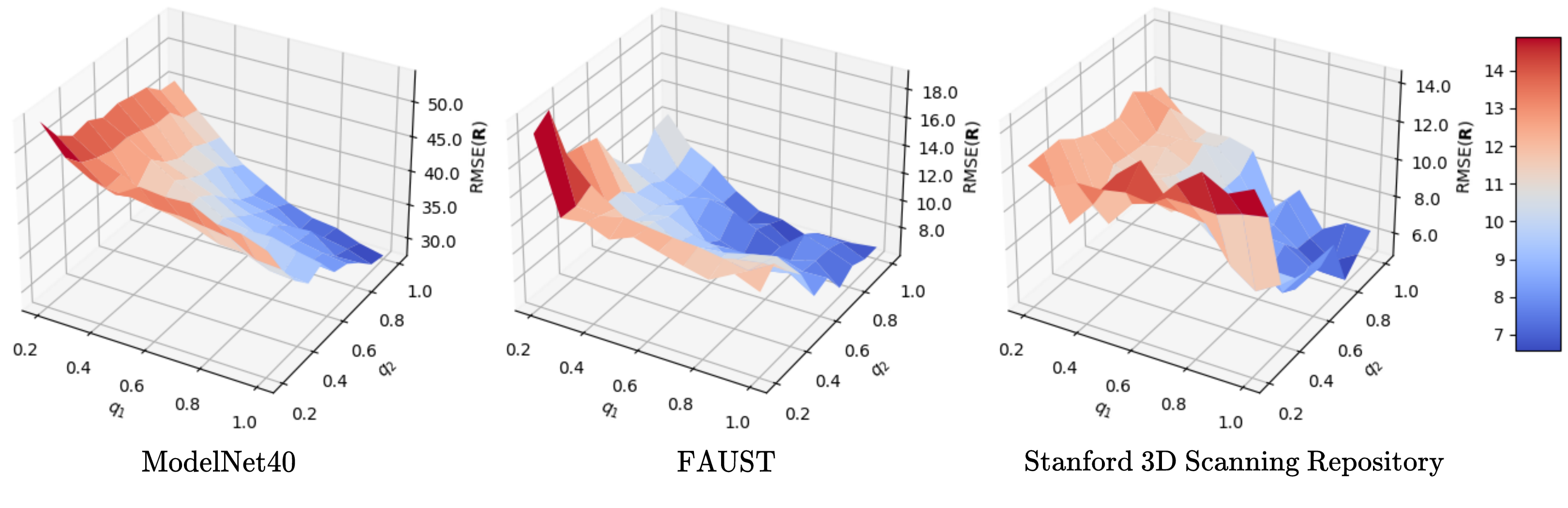}
    \caption{DeepUME performance (measured in RMSE($\vR$)) under Bernoulli noise model with respect to the probabilities $q_1$ and $q_2$ in all three data sets tested. The surfaces visualization implies that the sparseness of the sparsest point clouds is the dominant cause of error, rather then the difference between the clouds densities.}
    \label{fig:bernoulli}
    \end{minipage}
    
    \begin{minipage}{\linewidth}
    \centering
    \includegraphics[width=0.8\textwidth]{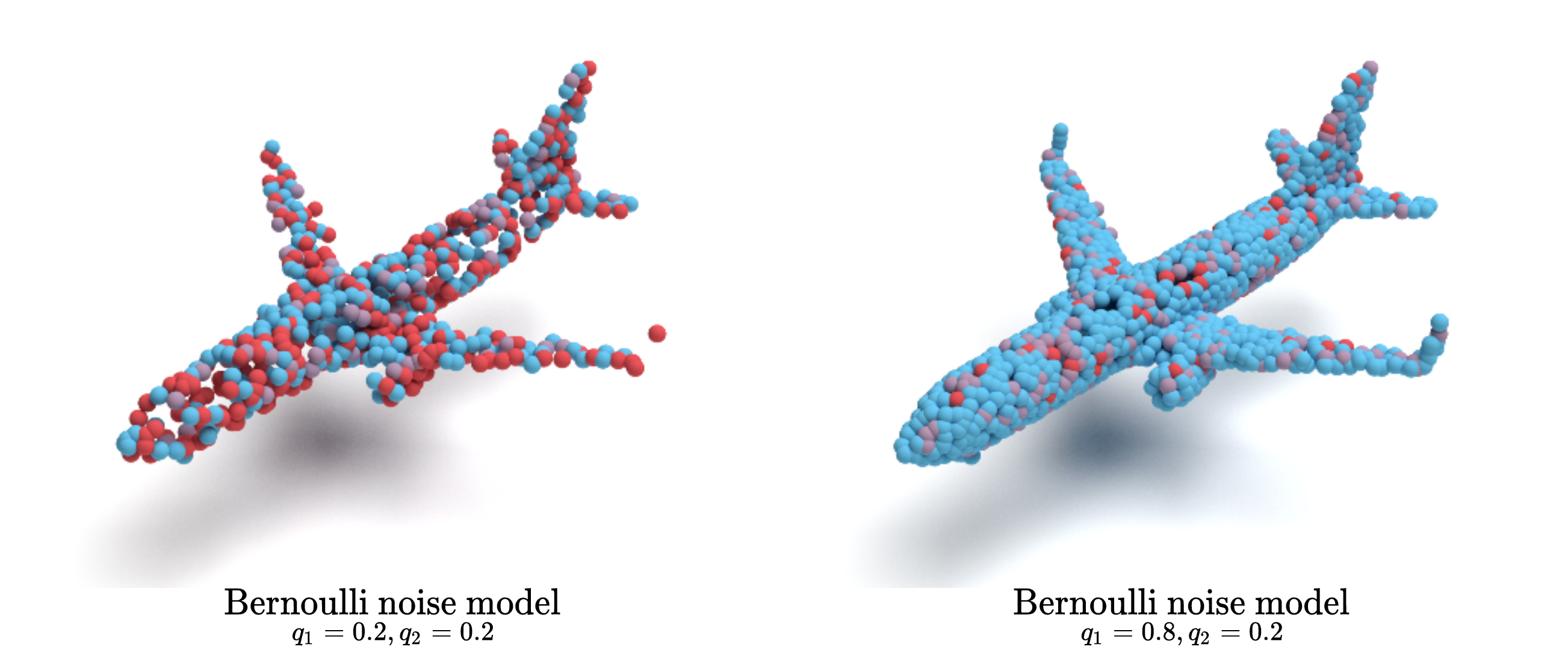}
    \caption{The effect of Bernoulli noise on a point cloud in extreme values of the probabilities $q_1$ and $q_2$. Red points denote $\cPC_1$, blue points denote $\cPC_2$ and purple points belong to both.}
    \label{fig:bernoulli_q1_q2}
    \end{minipage}
\end{figure}

The error surfaces depicted in Figure \ref{fig:bernoulli} describe the rotation RMSE with respect to the probabilities $q_1$ and $q_2$ in all three data sets tested.  As expected, the error increases as $q_1$ and $q_2$ decrease and the point clouds are made sparser. Note that in Figure \ref{fig:bernoulli}, in all the data sets considered, the rotation RMSE becomes large even when only one of the clouds is sparse (when the probability for keeping a point is small on one cloud and large on the other cloud). That is, the rotation RMSE does not increase much when we take the second cloud to be sparse as well. This might suggest that the sparseness of the sparsest point cloud is the dominant cause of error, rather then the difference between the clouds densities.  

For a visualization of the effect of Bernoulli noise on a point cloud in extreme values of the probabilities $q_1$ and $q_2$ we refer the reader to Figure \ref{fig:bernoulli_q1_q2}.

\subsection{Small rotation angles range} 
In our experiments, all DGCNN-based networks, except ours, provide poor registration. This is expected, as without a proper pre-processing procedure such networks fail to create features that are invariant under large rotations. The experimental results demonstrate that on the average, when large rotations are allowed, the registration results of the proposed method outperform the alternatives. 
However, for a more comprehensive overview, we also compare our performance separately for the case of  registration under rotation by small angles (up to $60^\circ$ about each axes). 

For a fair comparison, we use the pre-trained models released by the authors and test all the proposed methods  with rotation in the range $[0,60]$ degrees about each of the axes, for the noise free and zero-intersection noise models, in a similar manner to the experiments described in the main paper. We note that in the small rotations experiment, none of the tested methods encounters the ambiguity problem of ModelNet40, discussed in \ref{sec:ambiguity} and in section 5 in the main paper. The symmetry in the examples of ModelNet40 creates ambiguity where rotations by angles larger than $90^\circ$ are considered. Since our method is designed for arbitrary rotations, we do suffer from the ambiguity problem in ModelNet40. Hence, for generating a reliable comparison, we perform that experiment on the Stanford 3D Scanning Repository \cite{StanfordScanRep}, which does not contain ambiguous examples. 

From the results summarized in Table
\ref{tbl:truncated_angles}, we conclude that the method proposed in this paper outperforms all compared methods in all metrics examined (except for one case, where RGM method \cite{Fu2021RGM} achieves slightly smaller rotation RMSE). This shows that the proposed framework outperforms state-of-the-art methods not only on the average on the entire transformation range, but also for the small angles scenario. 

\begin{table}
\centering
\fontsize{7}{7}\selectfont
\setlength{\tabcolsep}{1.7pt} 
\renewcommand{\arraystretch}{1} 
\begin{tabular}{l cccc cccc}
\midrule
& 
\multicolumn{4}{c}{Noise free} & 
\multicolumn{4}{c}{Zero-intersection noise}\\

\cmidrule(lr){2-5} \cmidrule(lr){6-9} 

\textbf{Model} & \textbf{$d_\text{C}$}  & \textbf{$d_\text{H}$}  & \textbf{RMSE}(R) & \textbf{RMSE}(t) 
& \textbf{$d_\text{C}$}  & \textbf{$d_\text{H}$} & \textbf{RMSE}(R) & \textbf{RMSE}(t)\\
\midrule

UME \cite{efraim2019universal}  
       & <9e-05   & <4e-04   & 0.333   & \underline{<5e-06}
        & 0.030   & 0.269   & 42.559   & \underline{0.010}\\
        
ICP \cite{besl1992method}
        & 0.280   & 1.274   & 72.265   & 0.277
        & 0.267   & 1.256   & 71.532   & 0.276\\
\midrule
PointNetLK \cite{aoki2019pointnetlk}  
        & <7e-04   & 0.005   & 0.729   & 0.005
        & \underline{0.012}  & \underline{0.113}  & 11.504  & 0.095\\
        
DCP \cite{wang2019deep}  
        &<4e-04   &0.003   &5.739   &0.002
        &0.065   &0.479   &71.307   &0.011\\
        
DeepGMR \cite{yuan2020deepgmr}  
        & \underline{<4e-06}   & \underline{<2e-05}   & 0.110   & <6e-05
        & 0.294  & 0.270  & 33.128  & 0.015\\
        
RGM \cite{Fu2021RGM}  
        & 0.268   & 1.203   & \underline{0.025}   & <2e-04
        & 0.267   & 1.187  & \textbf{4.690}  & 0.032\\
       
\midrule
DeepUME (ours) 
        & \textbf{<1e-07}   & \textbf{<1e-07}   & <\textbf{8e-05}   & \textbf{<1e-07}
        &  \textbf{0.002}  & \textbf{0.118}  & \underline{5.159}  & \textbf{0.010}\\

\midrule
\end{tabular}
\caption{Free and zero-intersection noise models results on the unseen data set Stanford 3D Scanning Repository in small rotation angles regime of $[0, 60]$ degrees about each axis. Our method outperforms the competing techniques in the examined scenarios for all metrics, only except the RMSE($\vR$) in the zero-intersection noise model.
These results suggest that our framework improves the state-of-the-art methods not only on the average over the entire range, but also in the small angles scenario, where most of the methods were designed to be optimal.}
\label{tbl:truncated_angles}
\end{table}

\section{Ablation Study}\label{sec:ablations}
We conduct several ablation studies,  removing components of the proposed DeepUME and replacing each part with an alternative, to better evaluate our design.
The studies were tested for the Gaussian noise and zero-intersection noise models on the unseen FAUST \cite{bogo2014faust} data set,  in a similar manner to the experiments described in the main paper. The results of this section are summarized in Table \ref{tbl:ablation}.

\subsection{With or without invariant coordinates?}
We first try to evaluate whether the new transformation-invariant coordinates generated for the point cloud at the pre-processing phase, provide value over the original representation of the point cloud.
We therefore remove the pre-processig module (presented in Figure 2 in the paper)  and compare the resulting performance to that obtained using the full model. Table \ref{tbl:ablation} demonstrates that DeepUME performs consistently better with the inclusion of the pre-processig module.

\subsection{Coordinates joint-resampling or features joint-resampling?}
In our proposed framework, the Transformer layer executes a joint-resampling procedure in the coordinated space of $\R^3$. This is unlike other networks, where the joint-resampling process is executed in feature space. Applying the Transformer in the coordinate space has a notable computational advantage as in this case,  the embedding size of each point is significantly lower. In our our network the embedding size of each point on the coordinate space is $3$ while in the feature space it is $512$. Hence, performing a coordinate sampling allows for a dramatic decrease in computational complexity, both in the train and evaluation processes (twice faster). 
 We compare our framework with the two strategies - resampling in coordinate space and resampling in  features space. The results presented in Table \ref{tbl:ablation} show that the decrease in computational complexity does not cause a decrease in  registration accuracy.

\subsection{UME or MLP?}
While MLP (Multi Layer Perceptron) provides, in principle, a universal approximation, the UME integration into a DNN framework is designed to provide an accurate computation of a rigid motion under noisy sampling of point clouds. A natural question to ask is whether the UME parameter extraction may be replaced by a general learned module, such that registration with comparable accuracy is achieved. 
As expected, Table \ref{tbl:ablation} shows that the model performs better with the UME layer than a general MLP. 

\begin{table}
\centering
\fontsize{7}{7}\selectfont
\setlength{\tabcolsep}{1.7pt} 
\renewcommand{\arraystretch}{1} 
\begin{tabular}{l cccc cccc}
\midrule
& 
\multicolumn{4}{c}{Zero-intersection noise} & 
\multicolumn{4}{c}{Gaussian noise}\\

\cmidrule(lr){2-5} \cmidrule(lr){6-9} 

\textbf{Model} & \textbf{$d_\text{C}$}  & \textbf{$d_\text{H}$}  & \textbf{RMSE}(R) & \textbf{RMSE}(t) 
& \textbf{$d_\text{C}$}  & \textbf{$d_\text{H}$} & \textbf{RMSE}(R) & \textbf{RMSE}(t)\\
\midrule

No pre-processing
       & 0.094   & 0.913  & 83.506 & 0.140
       & 0.087   & 0.896  & 82.260 & 0.140\\

Features joint-sampling 
        & 0.007   & 0.082   & 14.913   & 0.027
        &  0.001  & 0.012   & 1.533  & 0.002\\
        
MLP instead of UME   
        & \underline{0.003}  & \underline{0.046}   & \underline{12.964}   & \underline{0.023}
        & \underline{0.001}  & \underline{0.011}  & \underline{1.488}  & \underline{0.003}\\
        
\midrule
DeepUME (full model) 
        & \textbf{0.002}   &\textbf{ 0.024}  & \textbf{8.630}  & \textbf{0.019}
        &\textbf{ 0.001 }  & \textbf{0.011 } & \textbf{1.069}   & \textbf{0.002}\\

\midrule
\end{tabular}
\caption{Ablation study results on the unseen data set FAUST \cite{bogo2014faust}. DeepUME full model achieves equal or better registration results in all tested scenarios and in all metrics.}
\label{tbl:ablation}
\end{table}

\section{Implementation Details}\label{sec:exp_sting}

\begin{figure}
\centering
\makebox[0pt]{\includegraphics[width=0.5\textwidth]{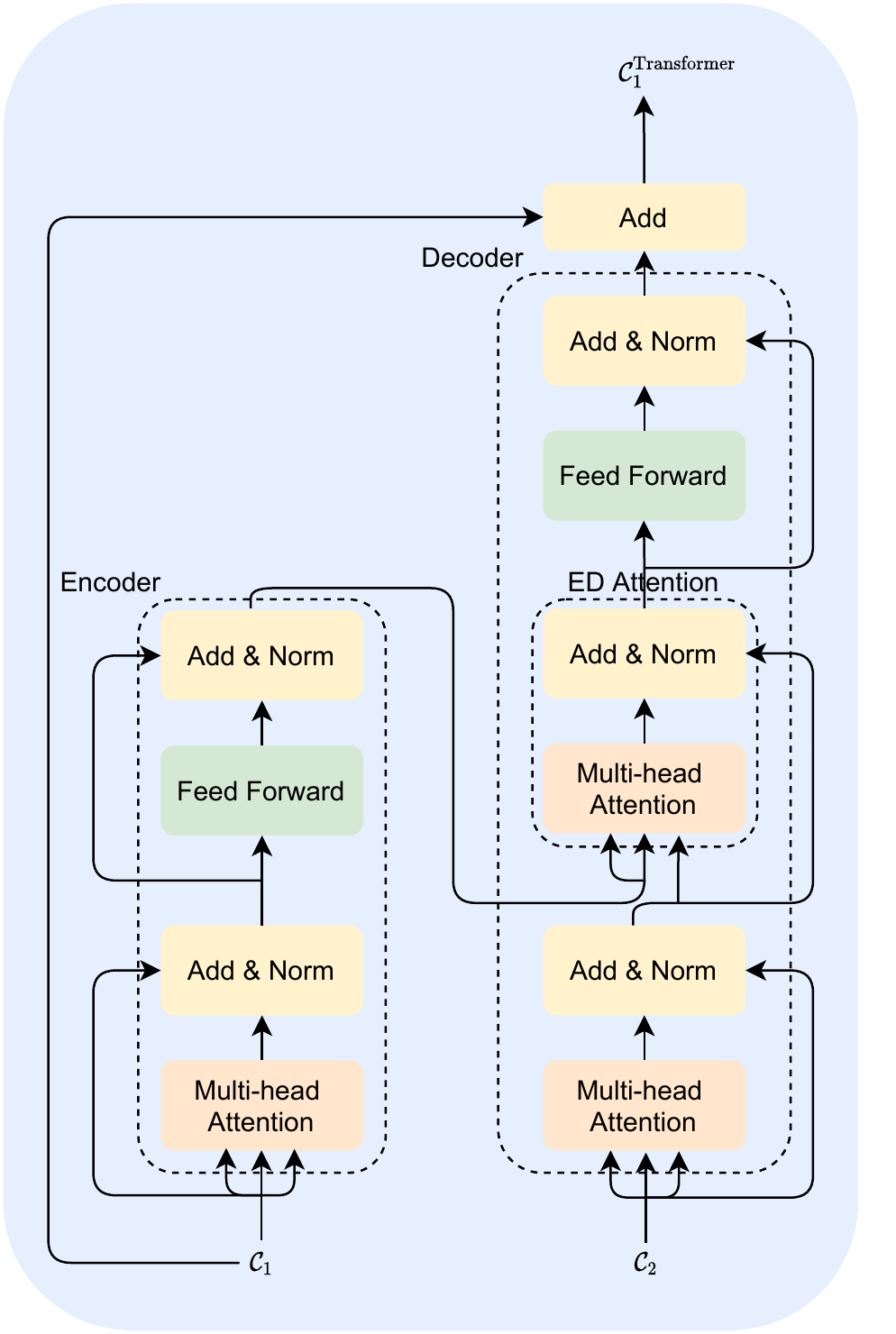}}
\caption{Transformer network architecture. ED stands for Encoder-Decoder.}
\label{fig:transformer}
\end{figure}
The architecture of DeepUME is shown in Figure 2 in the paper, and includes the Transformer and DGCNN learned modules.
The overall architecture of the Transformer \cite{vaswani2017attention} as used in this work, is depicted in Figure \ref{fig:transformer}. 
  
\subsection{Point clouds joint-resampling performed by a Transformer in the projected coordinate space}
Loosely speaking, the action of the Transformer is meant to improve performances in learned methods for various tasks, by jointly creating weighted sums that modify the original input. 
Formally, take $\cC_1$ and $\cC_2$ to be the point clouds generated by the module in \ref{sec:PCA}; these representations are computed independently of one another. Our attention model learns a function
\begin{equation}
\phi : \R^{N\times 3} \times \R^{M\times 3} \to \R^{N\times 3}
\end{equation}
that provides new coordinates for the point clouds
\begin{equation}
        \cC^\text{Transformer}_1=\cC_1 + \phi(\cC_1, \cC_2), \quad
        \cC^\text{Transformer}_2=\cC_2 + \phi(\cC_2, \cC_1).
\end{equation}
The objective of the map 
$\cC_1 \mapsto \cC^\text{Transformer}_1$ is to modify the features associated with the points in $\cC_1$ in a way that is knowledgeable of the structure of $\cC_2$; the map $\cC_2 \mapsto \cC^\text{Transformer}_2$ serves a symmetric role. The asymmetric function $\phi$ is given by a Transformer.

During the learning process,  the additive terms, $\phi(\cC_1, \cC_2)$ and $\phi(\cC_2, \cC_1)$,  change the point clouds, without distorting the shape. We therefore refer to this process as resampling.
A method for evaluating the similarity between the shapes of the Transformer input and output is to compute their Chamfer distance. We average the Chamfer distance between $\cC_i$ and $\cC^\text{Transformer}_i$ over the ModelNet40 data set, and get a distance of about $0.004$ (where all point clouds are scaled to the unit sphere). This shows that a different point cloud is obtained after the Transformer's action, however the small Chamfer distance between the input and output point clouds (relatively to the point clouds size) shows that this difference is due to a change of sampling points rather than a change of shape.   

\subsection{Optimization details} 
Adam \cite{kingma2014ba} is used to optimize the network parameters, with an initial learning rate of $0.001$. We divide the learning rate by $10$ at epochs $75, 150$ and $200$, training for a total of $250$ epochs.
As the modules of the pre-processing and the UME are of closed-form and non iterative or brute force, their impact on the computational time is negligible. 
DeepUME is implemented in Pyotrch \cite{NEURIPS2019_9015}, and its training times is about 11 hours long using an NVIDIA Quadro RTX6000 GPU.

\bibliography{supplement}